\documentclass[10pt,journal,compsoc]{IEEEtran}
\usepackage{epsfig}
\usepackage{graphicx}
\usepackage{amsmath}
\usepackage{booktabs}
\usepackage{amssymb}
\usepackage {multirow}
\usepackage{fixltx2e}
\usepackage{subfigure}
\usepackage{enumitem}

\newcommand{\vect}[1]{\mathbf{#1}}

\newcommand{\mr}[1]{\mathrm{#1}}

\def\eg{\textit{e.g.}~}
\def\ie{\textit{i.e.}~}

\def\etal{\textit{et al.}~}

\usepackage{tabularx}

\DeclareMathOperator*{\argmin}{arg\,min}

\newcommand{\tabincell}[2]{\begin{tabular}{@{}#1@{}}#2\end{tabular}}

\ifCLASSOPTIONcompsoc
  \usepackage[nocompress]{cite}
\else
  \usepackage{cite}
\fi
\ifCLASSINFOpdf
\else
\fi
\begin{document}
%
\title{Monocular Depth Estimation using Multi-Scale Continuous CRFs as Sequential Deep Networks}
%
%
%
%

\author{Dan Xu,~\IEEEmembership{Student Member,~IEEE,}
        Elisa Ricci,~\IEEEmembership{Member,~IEEE,}
        Wanli Ouyang,~\IEEEmembership{Senior Member,~IEEE,}
        Xiaogang Wang,~\IEEEmembership{Senior Member,~IEEE,}
        Nicu Sebe,~\IEEEmembership{Senior Member,~IEEE}
\IEEEcompsocitemizethanks{\IEEEcompsocthanksitem Dan Xu, Nicu Sebe are with the Department of Information Engineering and Computer Science, University of Trento, Trento, Italy. E-mail: \{dan.xu, niculae.sebe\}@unitn.it \protect
\IEEEcompsocthanksitem Elisa Ricci is with Fondazione Bruno Kessler. Email: eliricci@fbk.eu
\protect
\IEEEcompsocthanksitem Wanli Ouyang is with the School of Electrical and Information Engineering, The University of Sydney. Email: wanli.ouyang@sydney.edu.au \protect
\IEEEcompsocthanksitem Xiaogang Wang is with the Department of Electronic Engineering, The Chinese University of Hong Kong. Email: xgwang@ee.cuhk.edu.hk
}
\thanks{Manuscript received April 19, 2005; revised August 26, 2015.}}

%
%

\markboth{Journal of \LaTeX\ Class Files,~Vol.~14, No.~8, August~2015}%
{Shell \MakeLowercase{\textit{et al.}}: Bare Demo of IEEEtran.cls for Computer Society Journals}
%



\IEEEtitleabstractindextext{%
\begin{abstract}
Depth cues have been proved very useful in various computer vision and robotic tasks. This paper addresses the problem of monocular depth estimation from a single still image.
Inspired by the effectiveness of recent works on multi-scale convolutional neural networks (CNN), we propose a deep model which fuses complementary  
information derived from multiple CNN side outputs.
Different from previous methods using concatenation or weighted average schemes, the integration is obtained by means of 
continuous Conditional Random Fields (CRFs). In particular, we propose {two different variations}, 
one based on a cascade of multiple CRFs, the other on a unified graphical model. By designing a novel CNN implementation of mean-field updates for continuous CRFs, we show that both proposed models
can be regarded as sequential deep networks and that 
training can be performed end-to-end. 
Through an extensive experimental evaluation, we demonstrate the effectiveness of the proposed approach and establish new state of the art results for the monocular depth estimation task on three publicly available datasets, \ie NYUD-V2, Make3D and KITTI.
\end{abstract}

}

\maketitle

\IEEEdisplaynontitleabstractindextext

%
\IEEEpeerreviewmaketitle


%
%
%
%
%
%

\section{Introduction}
\IEEEPARstart{W}{hile} estimating the depth of a scene from
a single image is a natural ability for humans, devising computational models for 
accurately predicting depth information from RGB data is a challenging task.
Many attempts have been made to address this problem in the past. In particular, recent works have achieved 
remarkable performance thanks to powerful deep learning models \cite{eigen2015predicting,eigen2014depth,liu2015deep,porzi2017learning}. Assuming the 
availability of a large training set of RGB-depth pairs, 
monocular depth prediction from single images can be regarded as a pixel-level continuous regression problem and Convolutional Neural Network (CNN) architectures are typically employed.

In the last few years significant efforts have been made in the research community 
to improve the performance of CNN models for pixel-level prediction tasks (\eg semantic segmentation, contour detection).
Previous works have shown that, for depth estimation as well as for other pixel-level classification or regression problems, more accurate estimates can be obtained by combining information from multiple scales \cite{eigen2015predicting,xie2015holistically,chen2015attention, xu2017learningcvpr}. 
This can be achieved in different ways, \eg fusing feature maps corresponding to different network layers or 
designing an architecture with multiple inputs corresponding to images at different resolutions. 
Other works have demonstrated that, by adding a Conditional Random Field (CRF) in cascade to a convolutional neural architecture, the performance 
can be greatly enhanced and the CRF can be fully integrated within the deep model enabling end-to-end training 
with back-propagation \cite{zheng2015conditional}. However, these works mainly 
focus on pixel-level prediction problems in the discrete domain (\eg semantic segmentation).
While complementary, so far these strategies have been only considered in isolation and 
no previous works have exploited multi-scale information within a CRF inference framework. 

\begin{figure}[!t]
\centering
\includegraphics[width=3.4in]{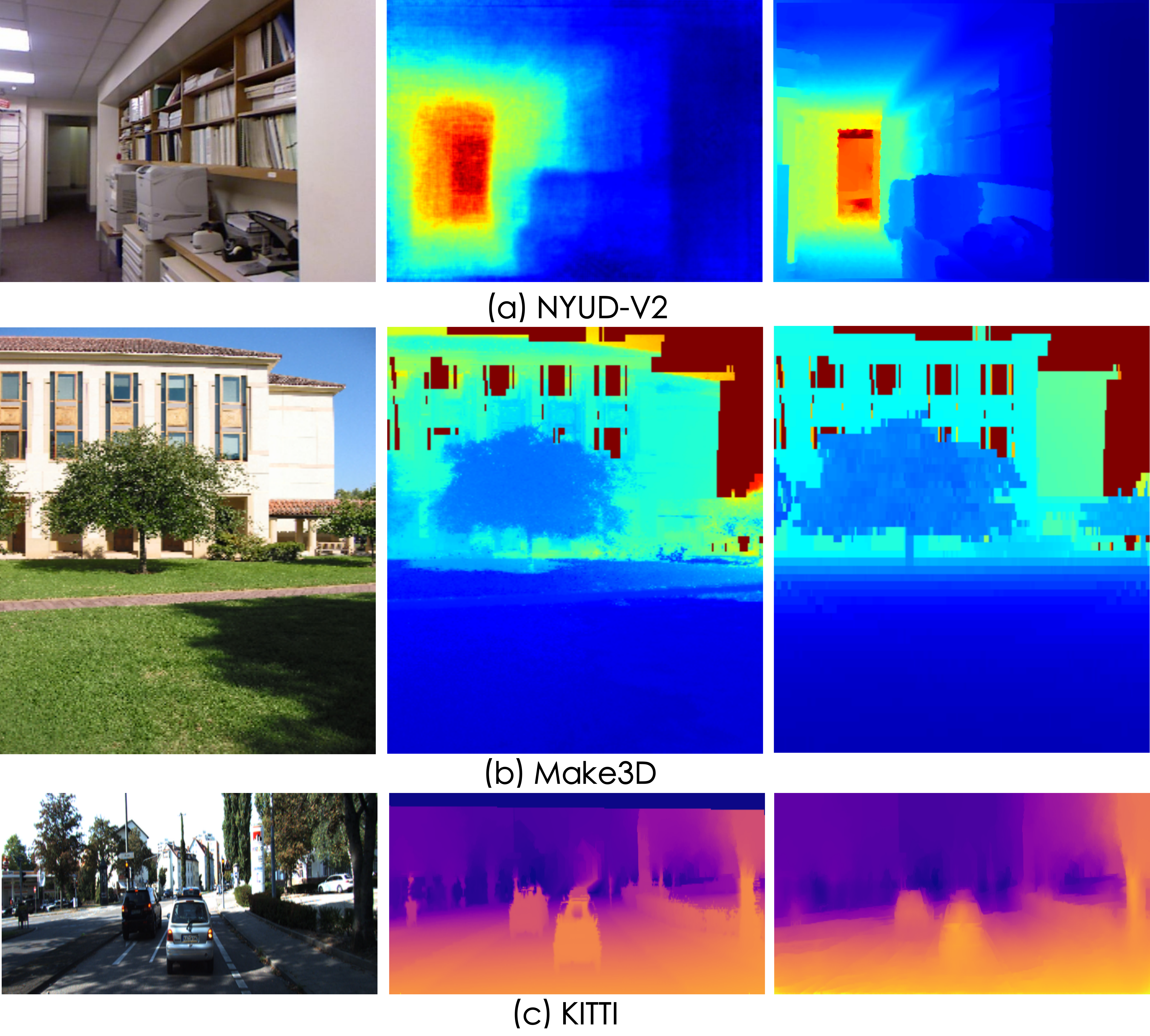}
\vspace{-10pt}
\caption{Monocular depth estimation results on three different benchmark datasets, \ie NYUD-V2 (the 1st row), Make3D (the 2nd row) and Kitti (the 3rd row), using the proposed multi-scale CRF model with a pre-trained CNN (\eg VGG Convolution-Deconvolution \cite{noh2015learning}). From left to right, each column is original RGB images, the recovered depth maps and the groundtruth, respectively. 
}
\vspace{-10pt}
\label{motivation}
\end{figure}

In this paper we argue that, benefiting from the flexibility and the representational power of graphical models, we
can optimally fuse representations derived from multiple CNN side-output layers using structured constraints, improving
performance over traditional multi-scale strategies.
By exploiting this idea, we introduce a novel framework
to estimate depth maps from single still images. 
{Opposite to previous work fusing multi-scale features by weighted averaging or concatenation, we propose to integrate 
multi-layer side-output information by designing a novel approach based on continuous CRFs. Specifically, we present two different methods. The first approach is based on a single multi-scale unified CRF model, while the other considers a cascade of scale-specific CRFs. We also show that, 
by introducing a common CNN implementation for mean-fields updates in continuous CRFs, both models are equivalent to 
sequential deep networks and an end-to-end approach can be devised for training. 
Through extensive experimental evaluation
we demonstrate that the proposed CRF-based approach produces more accurate depth maps than traditional multi-scale approaches for pixel-level prediction tasks \cite{hariharan2015hypercolumns,xie2015holistically}. Moreover, by performing experiments on the publicly available NYU Depth V2 \cite{silberman2012indoor}, Make3D~\cite{saxena2009make3d}  and KITTI~\cite{Geiger2013IJRR} datasets, we show that our approach is able to robustly reconstruct depth with good visual quality~(Fig.\ref{motivation}) and outperforms state of the art methods for the monocular depth estimation task.

\par This paper extends our earlier work~\cite{xu2017multi} through proposing and investigating different multi-scale connection structures for message passing, further enriching the related works, providing more approach details, and significantly expanding experimental results and analysis. To summarize, the contribution of this paper is threefold:
\begin{itemize}[leftmargin=*]
\item Firstly, we propose a novel approach for predicting depth maps from RGB inputs which exploits multi-scale estimations derived from CNN inner semantic layers by structurally fusing them within a unified CNN-CRF framework. 
\item Secondly, as the task of
pixel-level depth prediction implies inferring a set of continuous values, we show how
mean field (MF) updates can be implemented as sequential deep models, enabling end-to-end training of the whole network. We
believe that our MF implementation will be useful not only to researchers working on depth prediction, but also to
those interested in other problems involving continuous variables. Therefore, our code is made publicly available at https://github.com/danxuhk/ContinuousCRF-CNN.git.
\item Thirdly, our experiments demonstrate that
the proposed multi-scale CRF framework is superior to previous methods integrating information from different semantic network layers by combining multiple
losses \cite{xie2015holistically} or by adopting feature 
concatenations \cite{hariharan2015hypercolumns}. We also show that our approach outperforms state of the state of the art monocular depth estimation methods on 
public benchmarks and that
the proposed CRF-based models can be employed in combination with different pre-trained CNN architectures, consistently enhancing their performance. 
\end{itemize}
\par The remainder of this paper is organised as follows. We first introduce related work in Section~\ref{sec:releated}, and then the proposed
multi-scale CRF models for monocular depth estimation is presented in Section~\ref{sec:approach}. We further elaborate how the proposed models can be implemented as sequential neural network for end-to-end joint optimization in Section~\ref{sec:SequentialNN}. The experimental results and analysis are elaborated in Section~\ref{sec:exps}, and we conclude the paper in Section~\ref{sec:conclusion}.

\section{Related work}\label{sec:releated}
Our approach is built upon recent successes of deep CNN architectures for
image classification~\cite{krizhevsky2012imagenet, simonyan2014very, he2015deep} and fully convolutional networks for dense semantic image segmentation~\cite{long2015fully, noh2015learning}. We briefly introduce the most related works by organizing them into three main aspects, \ie monocular depth estimation, multi-scale CNN and dense pixel-level prediction via combination of CNN and CRFs. 

\par\textbf{Monocular depth estimation.} 
Previous approaches for depth estimation from single images can be grouped into three main categories: (i) methods operating on hand crafted features, (ii) methods based on graphical models and (iii) methods adopting deep convolutional neural networks.

\begin{figure*}[t]
\centering
\includegraphics[width=7.1in, height=3.3in]{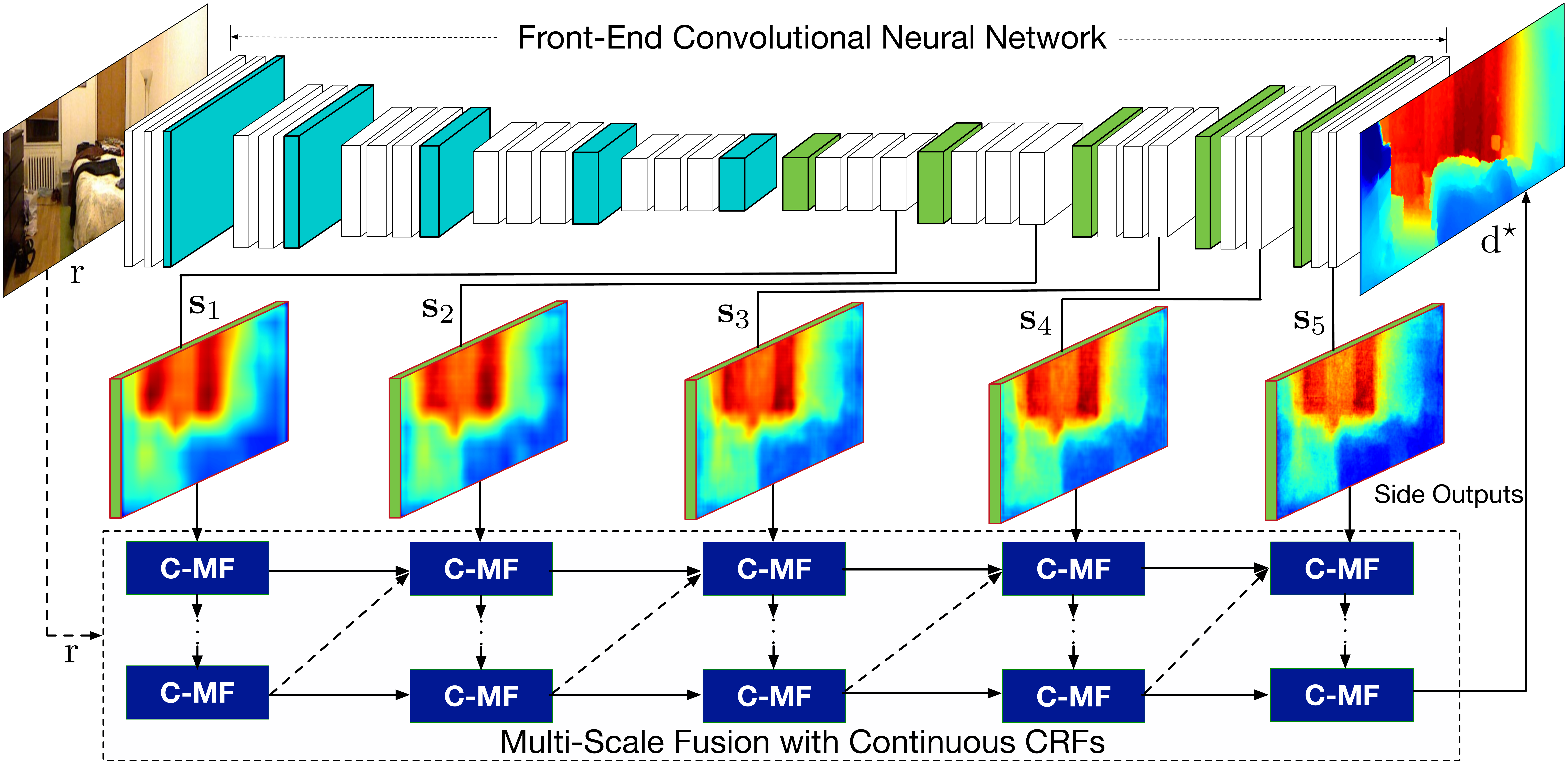} 
\vspace{-3pt}
\caption{Overview of the proposed deep architecture. Our model is composed of two main components: a front-end CNN and 
a fusion module. The fusion module uses continuous CRFs to integrate multiple side output maps of the front-end CNN. 
We consider two different CRFs-based multi-scale models and implement them as sequential deep networks by 
stacking several elementary blocks, the C-MF blocks. 
}
\label{framework}
\end{figure*}

\begin{figure*}[t]
\centering
\includegraphics[width=7.2in, height=0.75in]{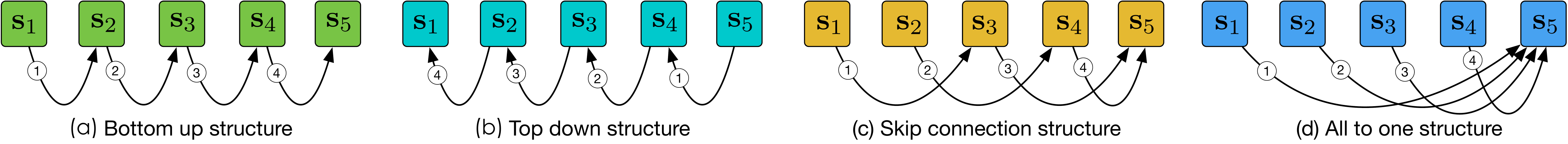} 
\vspace{-16pt}
\caption{Illustration of different multi-scale message passing structures for the integration of the multi-scale predictions $\mathbf{s}_{1}$ to $\mathbf{s}_{5}$ produced from the front-end convolutional network. The arrows represent the direction of the message passing, and the numbers in circles represent the order. The dashed line box in Fig.~\ref{framework} shows a bottom-up passing structure.
}
\vspace{-6pt}
\label{fig:messagepassingstructure}
\end{figure*}

Earlier works addressing the depth prediction task belong to the first category.
Hoiem \etal \cite{hoiem2005automatic, hoiem2005geometric} proposed photo pop-up, a 
fully automatic method for creating a basic 3D model from a single photograph by introducing an assumption of `ground-vertical' geometric structure. Karsch \etal \cite{karsch2014depth} developed Depth Transfer, a non parametric approach based on SIFT Flow, where the depth of an input image is reconstructed by transferring the depth of multiple similar images and then applying some
warping and optimizing procedures. Instead of directly recovering depth from appearance features, Liu \etal~\cite{liu2010single} explored using semantic scene segmentation results to guide the 3-D depth reconstruction. Similarly, Ladicky \etal \cite{ladicky2014pulling} also demonstrated the benefit of combining semantic object labels with depth features. However, the hand-crafted representations are not robust enough for this challenging problem. 

\par In the second category, some works exploited the flexibility of graphical models to reconstruct depth information.
For instance, Delage \etal \cite{delage2006dynamic} proposed a dynamic Bayesian
framework for recovering 3D information from indoor scenes. A discriminatively-trained 
multiscale Markov Random Fields (MRFs) were introduced in \cite{saxena20083, saxena2005learning}, in order to optimally fuse 
local and global features. Depth estimation was treated as an inference problem in a discrete-continuous CRF model in \cite{liu2014discrete}. However, these works did not 
employ deep networks.
\par More recent approaches for depth estimation are based on CNNs \cite{eigen2015predicting,liu2015deep,wang2015towards,roymonocular,laina2016deeper}.
For instance, Eigen \etal \cite{eigen2014depth} proposed a multi-scale approach for depth prediction,
considering two deep networks, one performing a coarse global prediction
based on the entire image, and the other refining predictions locally. This approach
was extended in \cite{eigen2015predicting} to handle multiple tasks (\eg semantic segmentation,
surface normal estimation). Wang \etal \cite{wang2015towards} introduced a CNN for joint depth estimation 
and semantic segmentation. The obtained estimates were further refined with Hierarchical CRFs. 
The most similar work to ours is \cite{liu2015deep}, where the 
representational power of deep CNN and continuous CRFs is jointly exploited for depth prediction.
However, the method proposed in \cite{liu2015deep} is based on superpixels and the information associated to multiple scales is not exploited in their graphical model. 
\par\textbf{Multi-Scale CNNs.} 
The problem of combining information from multiple scales has recently received considerable interest in various computer vision tasks. In \cite{xie2015holistically} a deeply supervised fully convolutional neural 
network was proposed for edge detection by weighted combination of multiple side outputs. Skip-layer networks, where the feature maps derived from different semantic layers of a primary front-end network are jointly considered in an output layer, have also become very popular
\cite{long2015fully,bertasius2015deepedge, chen2018group}. Other works considered multi-stream architectures, where multiple parallel networks receiving inputs at different scale are fused \cite{buyssens2012multiscale}. Cai~\etal~\cite{cai2016unified} proposed a multi-scale method via combining the predictions obtained from feature maps with different resolution for object detection. Dilated
convolutions (\eg \textit{dilation} or \textit{\`{a} trous}) have been also employed in different deep network models in order to 
aggregate multi-scale contextual information \cite{chen2014semantic}. However, in these works, the multi-scale representations or estimations are typically combined by using simple concatenation or weighted averaging operation. We are not aware of previous works exploring fusing deep multi-scale information within a CRF framework.

\par\textbf{Dense pixel-level prediction via combination of CNN and CRFs.} 
The combination of CNN and CRFs  has shown great usefulness for dense pixel-level structured prediction~\cite{schwing2015fully, knobelreiter2016end}. Some existing works utilize CRFs as a post processing module for further refining the predictions from the CNN~\cite{chen2016deeplab, papandreou2015weakly}. To benefit from end-to-end learning, Zhang \etal~\cite{zheng2015conditional} proposed a CRF-RNN model which jointly optimizes a front-end deep network with a discrete CRF for semantic image segmentation. Xu~\etal~\cite{xu2017learning} proposed an attention-gated deep CRF framework for pixel-level contour prediction. However, as far as we know, this work is a first attempt to combine multi-scale continuous CRFs with deep convolutional neural network for constructing a unified model for end-to-end monocular depth estimation.

\section{Multi-Scale CRF Models for Monocular Depth Estimation}\label{sec:approach}

In this section we introduce our deep model with the designed multi-scale continuous CRFs for monocular depth estimation from RGB images. We first formalize the problem of depth prediction and give a brief overview of the proposed approach. Then, we
describe two different variants of the proposed multi-scale model, one based on a cascade of CRFs and the other on a 
single multi-scale unified CRFs. 


\subsection{Problem Formulation and Overview}
Following previous works we formulate the task of depth prediction from monocular RGB input as the problem of learning a non-linear mapping 
$F:\mathcal{I} \rightarrow \mathcal{D}$ from the image space $\mathcal{I}$ to the output depth space $\mathcal{D}$. 
More formally, let $\mathcal{Q} = \{ (\vect{r}_i, \bar{\vect{d}}_i)\}_{i=1}^Q$ be a training set of $Q$ pairs, 
where $\vect{r}_i \in~\mathcal{I}$ denotes an input RGB image with $N$ pixels and $\bar{\vect{d}}_i \in\mathcal{D}$ 
represents its corresponding real-valued depth map. 

For learning $F$ we consider a deep model made of two main building blocks (Fig.~\ref{framework}). The first component is a CNN architecture with 
a set of intermediate side outputs $\mathcal{S}=\{ \vect{s}_{l} \}_{l=1}^{L}$, $\vect{s}_{l} \in R^N$, produced from $L$ different layers 
with a mapping function $f_s(\vect{r}; \mathbf{\Theta}, \boldsymbol{\theta}_{l}) \rightarrow \vect{s}_{l}$. 
For simplicity, we denote with $\mathbf{\Theta}$ the set of front-end network layer parameters and with $\boldsymbol{\theta}_{l}$ the parameters of 
the network branch producing the side output associated to the {$l$-th} layer (see Section \ref{setup} for details of our implementation). 
In the following we denote this network as the front-end CNN.

The second component of our model is a fusion block. As shown in previous works \cite{long2015fully,bertasius2015deepedge,xie2015holistically}, 
features generated from different 
CNN layers capture complementary information. The main idea behind the proposed fusion block
is to use CRFs to effectively integrate the side output maps of our front-end CNN 
for robust depth prediction. Our approach develops from the intuition that these representations 
can be combined within a sequential framework, 
\ie performing depth estimation at a certain scale
and then refining the obtained estimates in the subsequent level.
Specifically, we introduce and compare two different multi-scale models, both based on CRFs, and corresponding to two different versions of the
fusion block. 
The first model is based on a \textbf{single multi-scale unified CRFs}, which integrates information
available from different scales and simultaneously enforces smoothness constraints between the estimated depth values of
neighboring pixels and neighboring scales. The second model implements a \textbf{cascade of scale-specific CRFs}: at each scale $l$
a CRF is employed to recover the depth information from side output maps $\mathbf{s}_l$ and the 
outputs of each CRF model are
used as additional observations for the subsequent model.
In Section \ref{sec:HCRF} we describe the two models in details, while in Section \ref{mean-field}
we show how they can be implemented as sequential deep networks by 
stacking several elementary blocks. We call these blocks C-MF blocks, as they implement Mean Field updates for Continuous CRFs (Fig.~\ref{framework}).

\subsection{Multi-scale Fusion with Continuous CRFs}
We now elaborate the proposed CRF-based models for fusing multi-scale side-outputs derived from different semantic layers of the front-end deep convolutional neural networks. 

\begin{figure*}[!t]
\centering
\includegraphics[width=6.2in, height=3.1in]{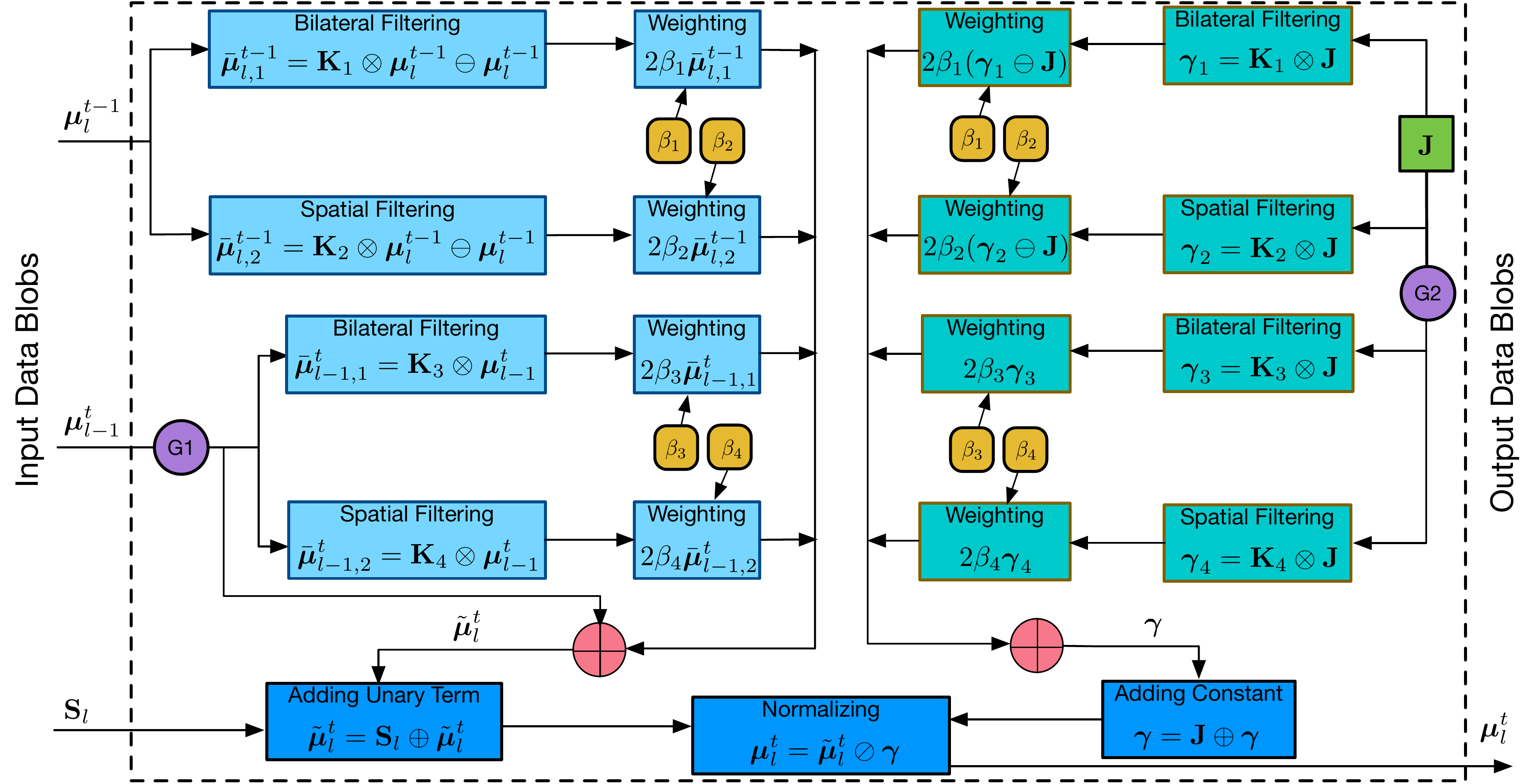} 
\vspace{-3pt}
\caption{Detailed computing flow graph of the proposed C-MF block. $\mathbf{J}$ represents a $W\times H$ matrix with all elements equal to one. 
The symbols $\oplus$, $\ominus$, $\oslash$ and $\otimes$ indicate element-wise addition, subtraction, division and Gaussian convolution operation, respectively. G1 and G2 represent two gate functions for controlling the computing flow.}
\vspace{-13pt}
\label{fig:HCRF-MF}
\end{figure*}

\subsubsection{Multi-Scale Unified CRF Model}
\label{sec:HCRF}
Given a vector $\hat{\mathbf{s}}$ with a dimension of $L \times N$ obtained by concatenating the side output 
{score maps} $\{\vect{s}_{1},\dots,\vect{s}_{L}\}$ and a vector $\vect{d}$ with a dimension of $L\times N$ expressing real-valued output variables, 
we define a CRF modeling the following conditional distribution:
\begin{equation}
\setlength{\abovedisplayskip}{2pt}
\setlength{\belowdisplayskip}{2pt}
  P(\vect{d}|\hat{\mathbf{s}}) = \frac{1}{Z(\hat{\mathbf{s}})} \exp \{-E(\vect{d},\hat{\mathbf{s}})\},
\end{equation}
where $Z(\hat{\mathbf{s}})=\int_{\vect{d}} \exp \{-E(\vect{d},\hat{\mathbf{s}})\} {d\vect{d}}$ is the partition function~\cite{lafferty2001conditional} acting as a normalization factor for probabilities.
The energy function is defined as:
\begin{equation}
\setlength{\abovedisplayskip}{2pt}
\setlength{\belowdisplayskip}{2pt}
E(\vect{d}, \hat{\mathbf{s}})  =  \sum_{i=1}^{N} \sum_{l=1}^{L} \phi(d_i^l, \hat{\mathbf{s}}) + \sum_{i, j} \sum_{l, k} 
 \psi(d_i^l, d_j^k),  \label{energy}
\end{equation}
and $d_i^l$ indicates the hidden variable associated to scale $l$ and pixel $i$.
The first term is the sum of quadratic unary terms defined as:
\begin{equation}
\setlength{\abovedisplayskip}{2pt}
\setlength{\belowdisplayskip}{2pt}
\phi(d_i^l, \hat{\mathbf{s}}) = \big (d_i^l - s_i^l \big )^2,
\end{equation}
where $s^l_i$ is the regressed depth value at pixel $i$ and scale $l$ obtained with $f_s(\vect{r}; \mathbf{\Theta}, \boldsymbol{\theta}_{l})$.
The second term is the sum of pairwise potentials describing the relationship 
between pairs of hidden variables $d_i^l$ and $d_j^k$ and is defined as follows:
\begin{equation}
\setlength{\abovedisplayskip}{2pt}
\setlength{\belowdisplayskip}{2pt}
\psi(d_i^l, d_j^k) =  \sum_{m=1}^{M} \beta_m w_m(i,j,l,k,\vect{r})(d_i^l-d_j^k)^2,
\end{equation}
where $w_m(i,j,l,k,\vect{r})$ is a weight which specifies the relationship between the estimated depth
of the pixels $i$ and $j$ at scale $l$ and $k$, respectively; $M$ is the number of kernels.

To perform inference we rely on the mean-field theory to approximate $P(\vect{d}|\hat{\mathbf{s}})$ with another distribution $Q(\vect{d}|\hat{\mathbf{s}})$, where $Q(\vect{d}|\hat{\mathbf{s}})=\prod_{i=1}^N\prod_{l=1}^L Q_{i,l}(d_i^l|\hat{\mathbf{s}})$, expressing a product of independent marginals. 
By minimizing the Kullback-Leibler divergence between the distribution of $P$ and $Q$, we obtain the solution of $Q$. As the log distribution $\log Q_{i,l}(d_i^l|\hat{\mathbf{s}})$ has a quadratic form w.r.t. $d_i^l$ and can be represented as Gaussian distribution, the following mean-field updates can be derived:
\begin{equation}
\setlength{\abovedisplayskip}{2pt}
\setlength{\belowdisplayskip}{2pt}
\gamma_{i,l} = 2 \big(1+2\displaystyle\sum_{m=1}^{M}\beta_m \sum_{k} \sum_{j, i} w_m(i,j,l,k,\vect{r})\big),
\label{sigma}
\end{equation}
\begin{equation}
\setlength{\abovedisplayskip}{2pt}
\setlength{\belowdisplayskip}{2pt}
\mu_{i,l} = \frac{2}{\gamma_{i,l}} \big(s_i^l + 2\sum_{m=1}^{M}\beta_m \sum_{k} \sum_{j, i} w_m(i,j,l,k,\vect{r}) \mu_{j,k} \big).
\label{mu}
\end{equation}
Here $\gamma_{i,l}$ and $\mu_{i,l}$ are the variance and mean of the distribution $Q_{i,l}$, respectively. 
\par To define the weights $w_m(i,j,l,k,\vect{r})$ we introduce the following assumptions. First, we assume that the estimated depth at scale $l$ only
depends on the depth estimated at previous scale. 
Second, for relating pixels at the same and at previous scale, we set
weights depending on $m$ kernel functions $K_m^{ij}$, which consists of Gaussian kernels with form of $\exp\big(-\frac{\|\mathbf{h}_i^m-\mathbf{h}_j^m\|^2_2}{2\theta_m^2}\big)$. Here,
$\vect{h}_i^m$ and $\vect{h}_j^m$ indicate some features 
derived from the input image $\vect{r}$ for pixels $i$ and $j$. $\theta_m$ are user-defined bandwidth parameters~\cite{koltun2011efficient}. 
Following previous works \cite{zheng2015conditional, koltun2011efficient}, we use pixel positions 
and color values as features, leading to two kernel functions, \ie a bilateral appearance kernel using both the pixel positions and the color value features and a spatial smoothness kernel using only the pixel positions features, 
for modeling dependencies of pixels at scale $l$ and other two for relating pixels at neighboring scales. 
Under these assumptions, the mean-field updates (\ref{sigma}) and (\ref{mu}) can be rewritten as:
\begin{equation}
\setlength{\abovedisplayskip}{2pt}
\setlength{\belowdisplayskip}{2pt}
\gamma_{i,l} = 2 \big(1+2 \sum_{m=1}^{2} \beta_m \sum_{j\neq i}  K_m^{ij} + 2\sum_{m=3}^{4} \beta_m 
\sum_{j, i} K_m^{ij}  \big),
\label{gamma2}
\end{equation}
\setlength{\abovedisplayskip}{2pt}
\setlength{\belowdisplayskip}{2pt}
\begin{equation}
\begin{aligned}
&\mu_{i,l} =& \frac{2}{\gamma_{i,l}} \big(s_i^l + 2\sum_{m=1}^{2}\beta_m \sum_{j \neq i} K_m^{ij} \mu_{j,l}, \\
&&+ 2\sum_{m=3}^{4} \beta_m \sum_{j, i} K_m^{ij} \mu_{j,l-1} \big).
\end{aligned}
\label{mu2}
\end{equation}
The parameters $\beta_m$ need to be learned during training. We will present the details of the parameter optimization in Section~\ref{sec:SequentialNN}. Given a new test image, the optimal $\tilde{\vect{d}}$ can be computed via maximizing the log conditional probability \cite{ristovski2013continuous}, \ie 
$\tilde{\vect{d}} = \mr{arg}\max_{\vect{d}} \log (Q(\vect{d}|\vect{S}))$,
where $\tilde{\vect{d}} = [\mu_{1,1},...,\mu_{N,L}]$ is a vector of the $L\times N$ mean values associated to $Q(\vect{d}|\hat{\mathbf{s}})$.
We take the estimated variables at the finest scale $L$ (\ie $\mu_{1,L},...,\mu_{N,L}$) as our predicted depth map $\mathbf{d}^\star$.

\subsubsection{Multi-Scale Cascade CRF Model}
\vspace{-3pt}
The cascade model is based on a set of $L$ CRF models, each one associated to a specific scale $l$, which are
progressively stacked such that the estimated depth at previous scale can be used as observations of the CRF model
in the following scale level.
Each CRF is used to compute the output vector $\vect{d}^l$ and it is 
constructed considering the side output representations $\vect{s}^l$ and the estimated depth at the previous
step $\tilde{\vect{d}}^{l-1}$ as observed variables, \ie $\vect{o}^l=[\vect{s}^l, \tilde{\vect{d}}^{l-1}]$.
The associated energy function of the CRF model is defined as:
\begin{equation}
\setlength{\abovedisplayskip}{2pt}
\setlength{\belowdisplayskip}{2pt}
E(\vect{d}^l, \vect{o}^l) = \sum_{i=1}^{N} \phi(d_i^l, \vect{o}^l) + \sum_{i \neq j} \psi(d_i^l, d_j^l).
\end{equation}
The unary and pairwise terms can be defined analogously to the above-introduced unified multi-scale model.
In particular the unary term, reflecting the similarity between the observation $o^i_l$ and the hidden depth value $d_i^l$, is:
\begin{equation}
\setlength{\abovedisplayskip}{2pt}
\setlength{\belowdisplayskip}{2pt}
\phi(y_i^l, \vect{o}^l) = \big (d_i^l - o_i^l \big )^2,
\end{equation}
where $o_i^l$ is obtained via combining the regressed depth from the side output $\vect{s}^l$ and 
the map $\vect{d}^{l-1}$ estimated by the CRF at previous scale. In our implementation we simply consider $o_i^l=s_i^l+\tilde{d}_i^{l-1}$, but other alternative strategies can be also considered.
The pairwise potentials, used to force neighboring pixels with similar appearance to have 
close depth values, are:
\begin{equation}
\setlength{\abovedisplayskip}{2pt}
\setlength{\belowdisplayskip}{2pt}
\psi(d_i^l, d_j^l) = \sum_{m=1}^{M}\beta_m K_m^{ij}(d_i^l-d_j^l)^2,
\end{equation}
where we consider $M=2$ Gaussian kernels, one for appearance features, and the other accounting for pixel positions.
Similar to the multi-scale CRF model, under mean-field approximation, the following updates can be derived:
\begin{equation}
\setlength{\abovedisplayskip}{2pt}
\setlength{\belowdisplayskip}{2pt}
\gamma_{i,l} = 2 \big(1+2 \sum_{m=1}^{M} \beta_m \sum_{j\neq i}  K_m^{ij}\big),
\label{gamma1}
\end{equation}
\begin{equation}
\begin{aligned}
\setlength{\abovedisplayskip}{2pt}
\setlength{\belowdisplayskip}{2pt}
&\mu_{i,l} =& \frac{2}{\gamma_{i,l}} \big(o_i^l + 2\sum_{m=1}^{M}\beta_m \sum_{j \neq i} K_m^{ij} \mu_{j,l}  \big).
\label{mu1}
\end{aligned}
\end{equation}
At the test time, we use the estimated depth variables corresponding to the cascade CRF model
of the finest scale $L$ as our final predicted depth map $\mathbf{d}^\star$.

\section{Multi-Scale Models as Sequential Deep Networks} \label{mean-field}\label{sec:SequentialNN}
In this section, we describe how the two proposed CRFs-based models can be implemented as sequential deep networks, enabling end-to-end training of our whole deep network model (the front-end CNN and the fusion module).
We first show how the mean-field iterations derived for the multi-scale and the cascade
models can be implemented by designing a common structure, the continuous mean-field updating (C-MF) block, {consisting into stack of a series of CNN operations}. Then, we present the resulting sequential network structures and details of the training phase for optimizing the whole deep network.

\begin{figure*} 
\centering 
\subfigure[The proposed multi-scale cascade CRF model as sequential neural network using the C-MF block.] { \label{fig:cascadeCRF} 
\includegraphics[width=5.3in, height=2.5in]{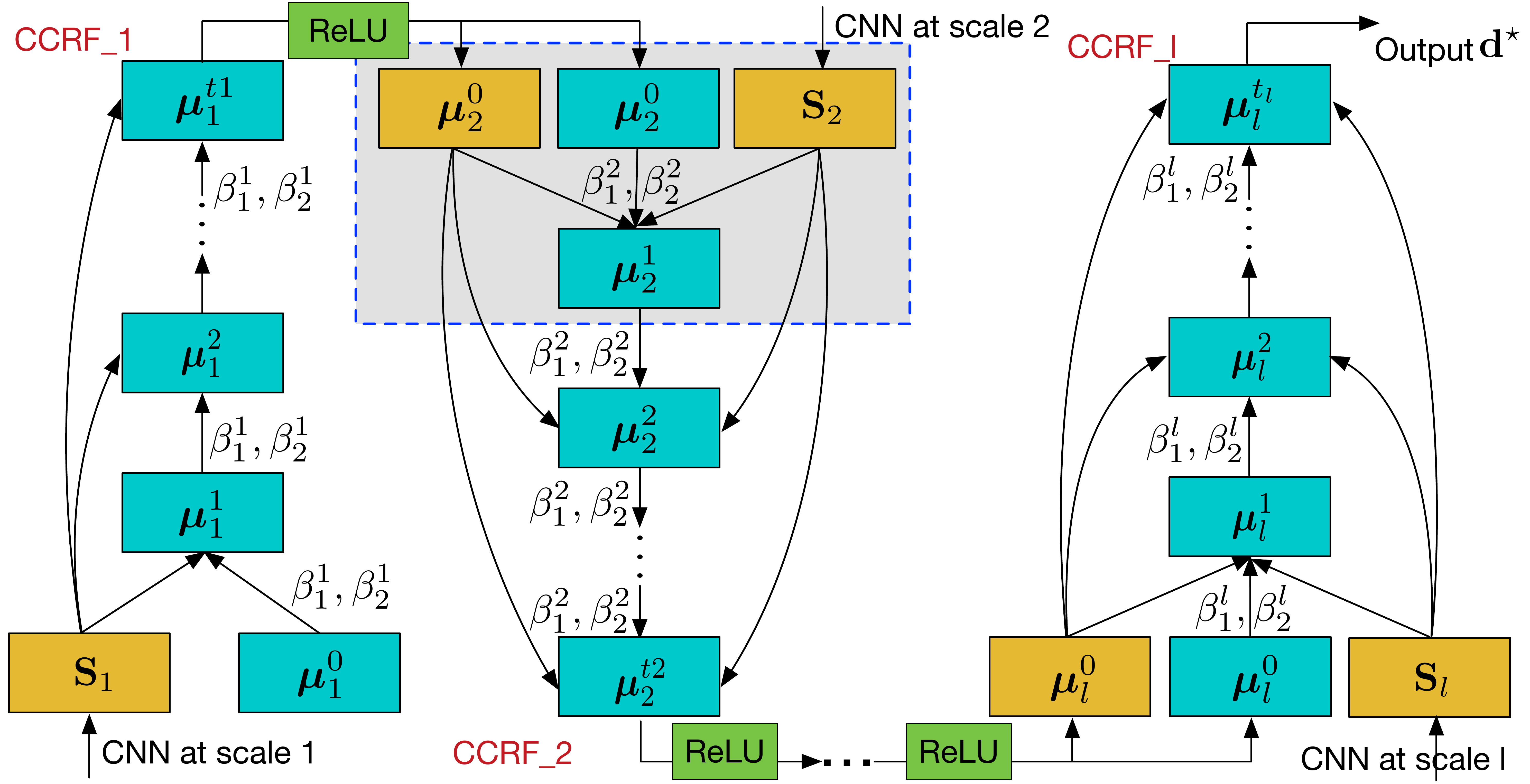} 
} 
\subfigure[The proposed multi-scale unified CRF model as sequential neural network using the C-MF block.] { \label{fig:multi-scaleCRF} 
\includegraphics[width=5.3in,height=2.5in]{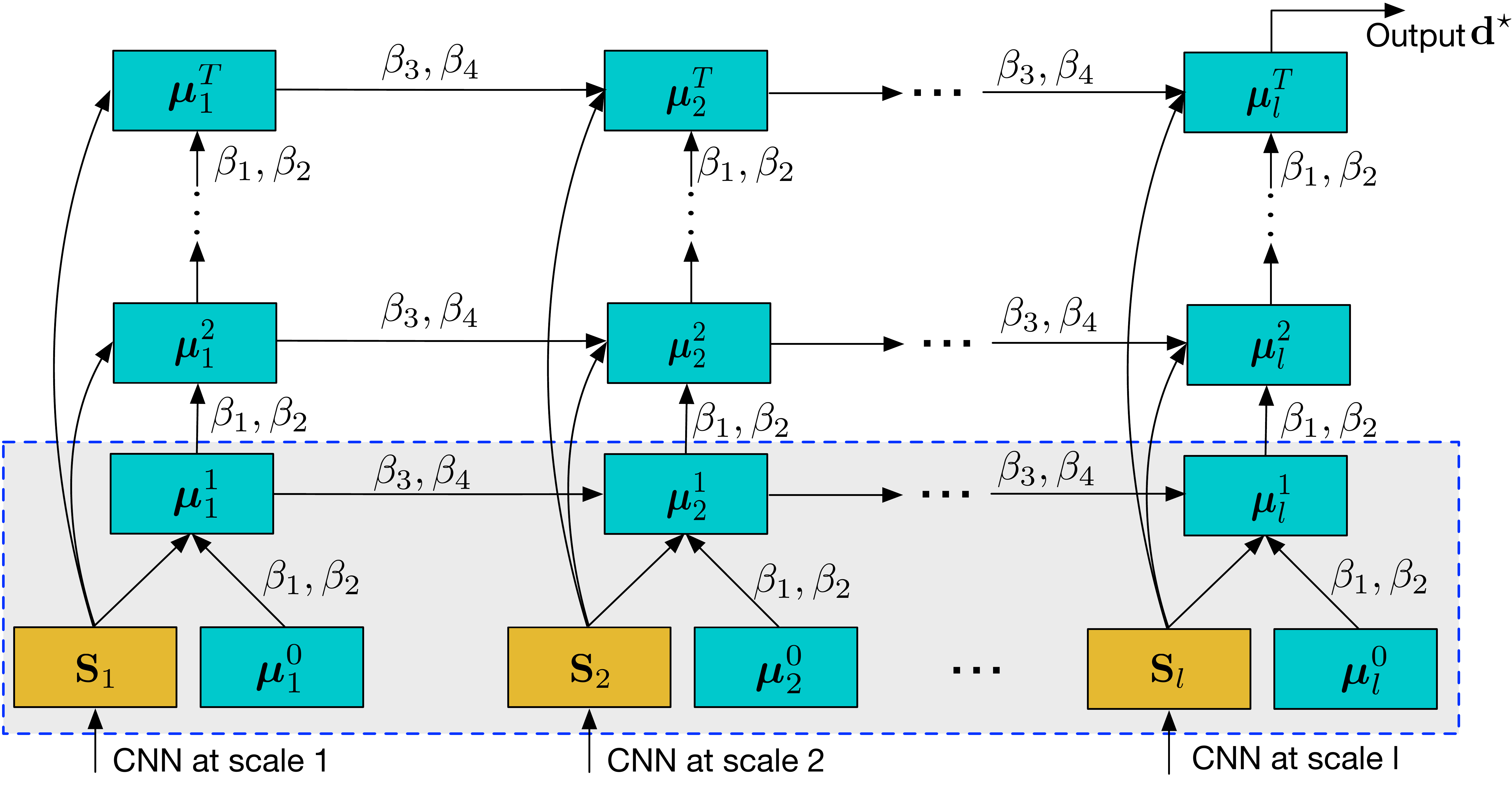} 
} 
\vspace{-10pt}
\caption{Description of the proposed two CRF models as sequential deep networks. The blue and yellow boxes indicate the
estimated variables and observations, respectively. The parameters ${\beta}_m$ are used for mean-field updates. 
As in the cascade model parameters are not shared among different CRFs, we use the notation 
${\beta}_1^l, {\beta}_2^l$ to denote parameters associated to the $l$-th scale. } 
\vspace{-6pt}
\label{fig} 
\end{figure*}

\subsection{C-MF: a Common {CNN} Implementation of Continuous Mean-Field Updating}
By analyzing the two proposed CRF models, we can observe that 
the mean-field updates derived for the cascade and for the multi-scale models
share common terms. As stated above, the main difference between the two is the way 
the estimated depth at previous scale is handled at the current scale. In the multi-scale CRFs, 
the relationship among neighboring scales is modeled in the hidden variable space, while in the cascade CRFs the depth estimated at previous scale acts as an observed variable.



Starting from this observation, in this section we show how the computation of Eq.~(\ref{mu2}) and Eq.~(\ref{mu1}) can be implemented 
with a common structure. 
Figure~\ref{fig:HCRF-MF} describes in details these computations. In the following, for the sake of
clarity, we introduce matrix representation. Let ${\mathbf{S}}_l \in \mathbb{R}^{W\times H}$ be the matrix obtained by rearranging 
the $N=W\times H$ pixels corresponding to the side output vector $\mathbf{s}_l$ and $\boldsymbol{\mu}_l^t \in \mathbb{R}^{W\times H}$ 
the matrix of the estimated output depth variables associated to scale $l$ and mean-field iteration $t$.
To implement the multi-scale model at each iteration $t$, $\boldsymbol{\mu}_l^{t-1}$ and $\boldsymbol{\mu}_{l-1}^t$ are convolved 
by two Gaussian kernels. Following \cite{koltun2011efficient}, we use a spatial and a bilateral kernel. As Gaussian convolutions 
represent the computational bottleneck (requiring a complexity of $\mathcal{O}(N^2)$) in the mean-field iterations, we adopt the permutohedral lattice implementation~\cite{adams2010fast} to approximate
the filter response calculation reducing the computational cost from quadratic to linear~\cite{ristovski2013continuous}.
The weighing of the parameters $\beta_m$ is performed as a convolution with a $1 \times 1$ kernel. Then, the outputs are combined and are added to the side-output maps $\mathbf{S}_l$. 
Finally, a normalization step follows, corresponding to the calculation of Eq.~(\ref{gamma2}). The normalization matrix 
$\boldsymbol{\gamma} \in \mathbb{R}^{W \times H}$ is also computed by considering convolutions with Gaussian kernels and 
weighting with parameters $\beta_m$. It is worth noting that the normalization step in our mean-field updates for 
continuous CRFs is substantially different from that of discrete CRFs in CRF-RNN~\cite{zheng2015conditional} based on a softmax function. 

In the cascade CRF model, differently from the multi-scale unified CRF model, $\boldsymbol{\mu}_{l-1}^t$ acts as an observed variable. To design a common C-MF block among the two models, we introduce two gate functions G1 and G2 (Fig.~\ref{fig:HCRF-MF}) controlling the computing flow and allowing to easily switch between the two approaches. Both gate functions accept a user-defined boolean parameter. In our setting, the value 1 corresponds to the multi-scale CRF and the value 0 corresponds to the cascade model. Specifically, if G1 is equal to 1, the gate function 
G1 passes $\boldsymbol{\mu}_{l-1}^{t}$ to the Gaussian filtering block, otherwise passes it to the element-wise addition block with the computed message. 
Similarly, G2 controls the computation of the normalization terms and switches between the computation of Eq.~(\ref{gamma2}) 
and Eq.~(\ref{gamma1}). In other words, if G2 equals to 0, then the Gaussian filtering and weighting operations for $\gamma_3$ and $\gamma_4$ are disabled. Importantly, for each step in the C-MF block we implement the calculation of error differentials for 
the back-propogation as in~\cite{zheng2015conditional}. 
\par There are two different types of CRF parameters to be learned, \ie the bandwidth parameters $\theta_m$ and the Gaussian-kernel weights $\beta_m$.  For optimizing these CRF parameters, similar to~\cite{koltun2011efficient}, the bandwidth values $\theta_m$ are pre-defined for simplifying the calculation, and we implement the backward differential computation for the weights of Gaussian kernels $\beta_m$. In this way $\beta_{m}$ are learned automatically with back-propagation.

\subsection{From Mean-Field Updates to Sequential Deep Networks}
Fig.~\ref{fig:HCRF-MF} illustrates the implementation of the proposed two CRF-based models using the designed C-MF block described above. 
In the figure, each blue-dashed box is associated to a mean-field iteration. 
The cascade model as shown in Fig.~\ref{fig:multi-scaleCRF} consists of $L$ single-scale CRFs. At the $l$-th scale, 
$t_l$ mean-field iterations are performed and then the estimated depth outputs are passed to another CRF model of the 
subsequent scale after a Rectified Linear Unit (ReLU) operation. The ReLU used here has two aspects of consideration: first the depth predictions should be always positive, and second we want to increase the nonlinearity of the sequential network for better mapping. To implement a single-scale CRF, we stack $t_l$ C-MF blocks and 
make them share the parameters, while we learn different parameters for different CRFs. For the multi-scale model, one full 
mean-field update involves $L$ scales simultaneously, obtained by combining $L$ C-MF blocks. We further stack $T$ 
iterations for learning and inference. The parameters corresponding to different scales and different 
mean-field iterations are shared. In this way, by using the common C-MF layer, we implement the two proposed multi-scale continuous CRFs models as 
deep sequential networks enabling end-to-end training with the front-end network.
\subsection{Multi-Scale Message Passing Structures}\label{sec:messagepassingstructure}
The proposed work aims at multi-scale structured fusion and prediction, the connection structure between the different multi-scale predictions for message passing plays an important role in the performance. In this section, we thus propose and investigate different message passing structures. Fig.~\ref{fig:messagepassingstructure} illustrates several structures include top down structure, skip-connection structure and all to one structure. The top down structure is similar to the bottom up structure depicted in Fig.~\ref{framework}, which gradually refines the score maps from coarse to fine. The skip connection structure aims at utilizing more complementary information via skipping scales. The all to one structure uses all the other scales to refine the finest scale. Since all the message passing structures involve two scales at each time, we are able to build all these proposed connection structures by using the proposed aforementioned neural-network implemented C-MF block. The experimental investigation of these structures is illustrated in the experimental part.

\subsection{Optimization of The Whole Network}\label{sec:optimization}
We train the whole network using a two phase scheme. In the first phase (pretraining), the parameters of the base front-end network $\mathbf{\Theta}$ and the parameters of the side-output generation sub-branch networks $\boldsymbol{\vartheta} = \{\boldsymbol{\theta}_l\}_{l=1}^L$ are learned
by minimizing the sum of $L$ distinct side losses as in \cite{xie2015holistically}, corresponding to $L$ side 
outputs. We define the optimization objective using a square loss over $Q$ training samples as follows:
\begin{equation}
\{\mathbf{\Theta}^*, \boldsymbol{\vartheta}^* \}=\argmin_{\mathbf{\Theta}, \boldsymbol{\theta}_{l}}\sum_{l=1}^L  \sum_{i=1}^{Q}\|f_s(\vect{r}_i; \mathbf{\Theta}, \boldsymbol{\theta}_{l}) - \vect{\tilde{d}}_i\|_2^2,
 \end{equation} 
 where $\vect{\tilde{d}}_i$ denotes the $i$-th ground-truth sample. 
In the second phase (fine tuning), we initialize the front-end network with the learned parameters $\{\mathbf{\Theta}^*, \boldsymbol{\vartheta}^*\}$ in the first phase, and 
jointly fine-tune with the proposed multi-scale CRF models to compute the optimal value of the 
parameters $\mathbf{\Theta}$, $\boldsymbol{\vartheta}$ and $\boldsymbol{\beta}$, with 
$\boldsymbol{\beta}=\{\beta_m\}_{m=1}^M$. The entire network is learned with Stochastic Gradient Descent (SGD) 
by minimizing a square loss 
\begin{equation}
\{{\mathbf{\Theta}^*, \boldsymbol{\vartheta}^*, \boldsymbol{\beta}^*}\}=\argmin_{\mathbf{\Theta}, \boldsymbol{\vartheta}, \boldsymbol{\beta}}\sum_{i=1}^{Q}\|F(\vect{r}_i; \mathbf{\Theta},\boldsymbol{\vartheta},\boldsymbol{\beta}) - \vect{\tilde{d}}_i\|_2^2.
\end{equation}
When the whole network optimization is finished, the test can be performed end-to-end, \ie given a test RGB image as input the network directly outputs an estimated depth map.

\section{Experiments}\label{sec:exps}
\begin{table}[t]
\centering
\caption{The parameter details of the sub-network for generating the side output from the last-scale convolutional block of ResNet-50.}
\vspace{-3pt}
\begin{tabular}{l|ccc}
\toprule
Name   &   conv\_s5\_1  &  deconv\_s5\_1  & deconv\_s5\_2  \\ 
\hline
Type      &   conv   & deconv & deconv \\
Kernel   & $3\times 3 \times 1024$ & $4 \times 4 \times 512 $ & $4 \times 4 \times 256$ \\
Stride, Padding  &  1, 1 & 2, 1 & 2, 1 \\
Activation & ReLU & ReLU  & ReLU\\
\hline
Name   &   deconv\_s5\_3  &  deconv\_s5\_4  & pred\\\hline
Type     &   deconv  & deconv & deconv \& crop \\
Kernel  &  $4 \times 4 \times 128 $ & $4 \times 4 \times 64 $ &  $ 4 \times 4 \times 1 $ \\
Stride, Padding & 2, 1 & 2, 1 & 2, 1 \\
Activation & ReLU &  ReLU & - \\
            
\bottomrule
\end{tabular}
\label{tab:side-outputs}
\vspace{-10pt}
\end{table}

\begin{figure*}[!t]
\centering
\includegraphics[width=1.01\linewidth]{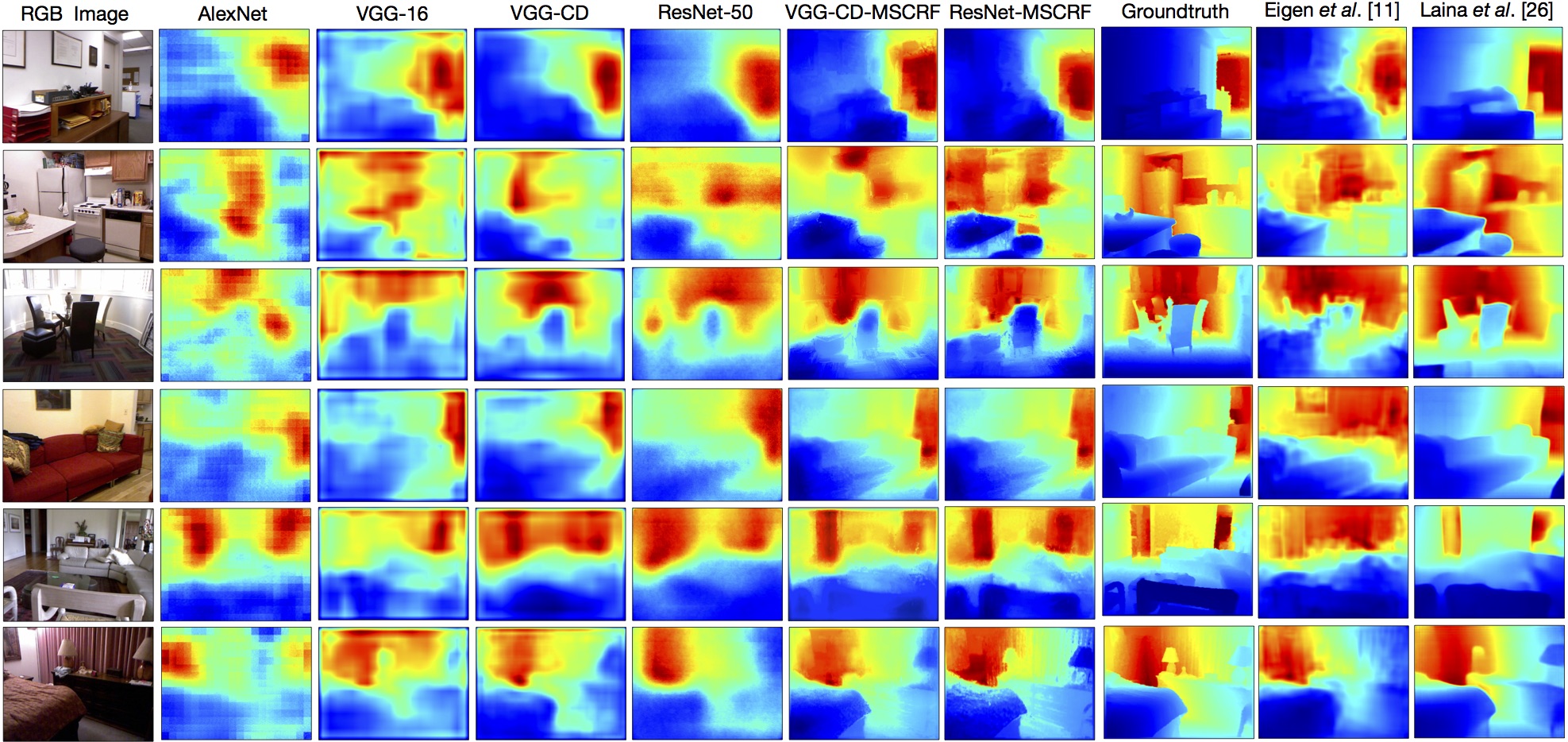} 
\caption{Examples of qualitative depth prediction results of different methods on the NYU v2 test dataset. Different front-end deep network architectures are investigated. VGG-CD-MSCRF and ResNet-MSCRF represent our approach with the proposed multi-scale continuous CRF model plugged on VGG-CD and ResNet-50 network respectively.}
\label{nyu2examples}
\vspace{-10pt}
\end{figure*}

To demonstrate the effectiveness of the proposed multi-scale CRF models for monocular depth prediction, we performed experiments on three publicly available datasets: the NYU Depth V2 ~\cite{silberman2012indoor}, the Make3D~\cite{saxena2005learning} and the KITTI~\cite{Geiger2013IJRR} datasets. In the
following we first describe the experimental setup and the implementation details, and then present the experimental results and analysis. 

\subsection{Experimental Setup}
\label{setup}
\subsubsection{Datasets}
The \textbf{NYU Depth V2} dataset~\cite{silberman2012indoor} contains 120K unique pairs of RGB and depth images captured with a 
Microsoft Kinect. The datasets consists of 249 scenes for training and 215 scenes for testing. The images
have a resolution of $640 \times 480$. To speed up the training phase, following previous works~\cite{liu2015deep,zhuo2015indoor}
we consider only a small subset of images. This subset has 1449 aligned RGB-depth pairs:
795 pairs are used for training, 654 for testing. Following~\cite{eigen2014depth}, we perform data augmentation for 
the training samples. The RGB and depth images are scaled with a ratio $\rho \in \{1, 1.2, 1.5\}$ and the depths are divided by $\rho$. Additionally,
we horizontally flip all the samples and randomly crop them to $320 \times 240$ pixels. The data augmentation phase produces 
4770 training pairs in total. 

The \textbf{Make3D} dataset~\cite{saxena2005learning} contains 534 RGB-depth pairs, split into 400 pairs for training and 134 for testing. 
We resize all the images to a resolution of $460 \times 345$ as done in~\cite{liu2014discrete} to preserve the aspect ratio 
of the original images. We adopted the same data augmentation scheme used for NYU Depth V2 dataset but,
for $\rho=\{1.2,1.5\}$ we randomly generate two samples each via cropping, obtaining 4K training samples. 

%

The \textbf{KITTI} dataset~\cite{Geiger2013IJRR} is built for various computer vision tasks within the context of autonomous driving, which contains depth videos captured through a LiDAR sensor deployed on a driving vehicle. For the training and testing split, we follow the protocol made by Eigen~\etal~\cite{eigen2014depth} for a better comparison with existing works. Specifically, 61 scenes are selected from the raw data. Total 22,600 images from 32 scenes are used for training, and 697 images from the other 29 scenes are used for testing. Following~\cite{garg2016unsupervised}, the ground-truth depth maps are generated by reprojecting the 3D points collected from velodyne laser into the left monocular camera. The resolution of RGB images are reduced half from original $1224\times 368$ for training and testing.  

%

\subsubsection{Evaluation Metrics}
Following previous works~\cite{eigen2015predicting,eigen2014depth,wang2015towards}, we adopt the following evaluation metrics 
to quantitatively assess the performance of our depth prediction model. Specifically, we consider: 
\begin{itemize}
\item mean relative error (rel): 
\( \frac{1}{P}\sum_{i=1}^P\frac{|\tilde{d}_i - d_i^\star|}{d_i^\star} \); 
\item root mean squared error (rms): 
\( \sqrt{\frac{1}{P}\sum_{i=1}^P(\tilde{d}_i - d_i^\star)^2} \);
\item mean log10 error (log10): \\
\( \frac{1}{P}\sum_{i=1}^P \Vert \log_{10}(\tilde{d}_i) - \log_{10}(d_i^\star) \Vert \);
\item scale invariant rms log error as used in~\cite{eigen2014depth}, rms(sc-inv.);
\item accuracy with threshold $t$: percentage (\%) of $d_i^\star$, \\subject to $\max (\frac{d_i^\star}{\tilde{d}_i}, \frac{\tilde{d}_i}{d_i^\star}) = 
\delta < t~(t \in [1.25, 1.25^2, 1.25^3])$.
\end{itemize}
Where $\tilde{d}_i$ and $d_i^\star$ is the ground-truth depth and the estimated depth at pixel $i$ respectively; $P$ is the total number of pixels of the test images. 

\subsection{Implementation Details}
We implemented the proposed deep model using the popular Caffe framework [15] on a single Nvidia Tesla K80 GPU with 12 GB memory. More details on the front-end CNN architectures, the generation of multi-scale side outputs and the parameter settings are elaborated as follows.

\begin{table*}
\centering
\caption{Quantitative performance comparison of different front-end deep network architectures and the proposed two multi-scale CRF models associated with the pretrained front-end networks on the NYU Depth V2 dataset. }
\vspace{-3pt}

\setlength\tabcolsep{8pt}
\resizebox{0.86\linewidth}{!} {
\begin{tabular}{l|ccc|ccc}
\toprule
\multirow{2}{*}{\tabincell{c}{Network  Architecture}} & \multicolumn{3}{c|}{\tabincell{c}{Error \\ (lower is better)}} & \multicolumn{3}{c}{\tabincell{c}{Accuracy \\ (higher is better)}} \\\cline{2-7}
                                      & rel & log10 & rms & $\delta < 1.25$ & $\delta < 1.25^2$ & $\delta < 1.25^3$ \\\midrule
AlexNet (pretrain)       & 0.265 &0.120&0.945 & 0.544 & 0.835 & 0.948 \\
VGG-16 (pretrain)                  &0.228  &0.104&0.836& 0.596 & 0.863 & 0.954 \\
VGG-ED (pretrain)               & 0.208 & 0.089 &0.788& 0.645 & 0.906 & 0.978\\
VGG-CD (pretrain)               & 0.203 & 0.087&0.774&  0.652 & 0.909& 0.979 \\
ResNet-50 (pretrain)             & 0.168& 0.072& 0.701& 0.741 & 0.932 & 0.981\\ 
\midrule
AlexNet + cascade-CRFs              & 0.231 &  0.105     & 0.868 &0.591 & 0.859 & 0.952 \\
VGG-16 + cascade-CRFs               &0.193    &0.092& 0.792    &   0.636     &     0.896    &    0.972     \\
VGG-ED + cascade-CRFs             &0.173    &0.073& 0.685    &   0.693     &     0.921    &    0.981     \\
VGG-CD + cascade-CRFs             & 0.169  & 0.071& 0.673    &   0.698     &     0.923    &  0.981 \\
ResNet-50 + cascade-CRFs            & \textbf{0.143} &\textbf{0.065}& \textbf{0.613}&\textbf{0.789} & \textbf{0.946} & \textbf{0.984} \\
\bottomrule                            
\end{tabular}
}
\label{cnn_arch}
\vspace{-5pt}
\end{table*}

\subsubsection{Front-end CNN Architectures}
To study the influence of the frond-end CNN, we consider several network architectures including: (i) AlexNet~\cite{krizhevsky2012imagenet}, (ii) VGG-16~\cite{simonyan2014very}, 
(iii) a fully convolutional encoder-decoder network derived from VGG-16, referred as VGG-ED~\cite{badrinarayanan2015segnet}, 
(iv) a Convolution-Deconvolution network based on VGG-16, referred as VGG-CD~\cite{noh2015learning}, and (v) ResNet-50~\cite{he2015deep}. For AlexNet, VGG-16 and ResNet-50, we obtain the side outputs from the last semantic convolutional layer of different convolutional blocks, in which each the layer produces feature maps with the same shape. The scheme utilized for the generation will be introduced in the next section. The number of side outputs considered in our experiments is 5, 5 and 4 for AlexNet, VGG-16 and ResNet-50, respectively. As VGG-ED and VGG-CD have been widely used for dense pixel-level prediction tasks, we also investigate them in the experimental analysis. Both VGG-ED and VGG-CD have a symmetric network structure, and five side outputs are then generated from the different blocks of the decoder or the deconvolutional network part.


\begin{table*}
\centering
\caption{Quantitative baseline comparison with different multi-scale fusion schemes, and with the continuous CRF as a post-processing module on the NYU Depth V2 dataset. The number of scales is investigated for both multi-scale models with a bottom up message passing structure. }
\vspace{-3pt}
\setlength\tabcolsep{10pt}
\resizebox{0.86\textwidth}{!} {
\begin{tabular}{l|ccc|ccc}
\toprule
\multirow{2}{*}{Method} & \multicolumn{3}{c|}{\tabincell{c}{Error \\ (lower is better)}} & \multicolumn{3}{c}{\tabincell{c}{Accuracy \\(higher is better)}} \\\cline{2-7}
                                      & rel & log10 & rms & $\delta < 1.25$ & $\delta < 1.25^2$ & $\delta < 1.25^3$ \\\midrule
HED~\cite{xie2015holistically}      & 0.185 & 0.077    &    0.723   & 0.678& 0.918&  0.980\\
Hypercolumn~\cite{hariharan2015hypercolumns}      &0.189 & 0.080 & 0.730 & 0.667& 0.911&  0.978\\
C-CRF             & 0.193& 0.082& 0.742& 0.662 & 0.909 & 0.976\\ \midrule
Ours (single-scale)                & 0.187     &    0.079   &    0.727  & 0.674& 0.916&  0.980 \\
Ours - cascade (3-scale)              &0.176   &    0.074    &    0.695  & 0.689 & 0.920&  0.980     \\
Ours - cascade (5-scale)            &0.169  & 0.071& 0.673  & 0.698 & 0.923 &  \textbf{0.981}     \\
Ours - unified (3-scale)           &  0.172 & 0.072   &  0.683   &  0.691   &  0.922   &  \textbf{0.981} \\ 
Ours - unified (5-scale)           &  \textbf{0.163} &  \textbf{0.069} &   \textbf{0.655}  & \textbf{0.706 }  &\textbf{0.925}    &   \textbf{0.981}\\ 
\bottomrule                            
\end{tabular}
}
\label{comparison}
\vspace{-5pt}
\end{table*}

\begin{table*}
\centering
\caption{Quantitative performance evaluation of different message passing structures for the cascade CRF model via building the sequential deep network with the proposed C-MF block on the NYU Depth V2 dataset. }
\vspace{-3pt}
\footnotesize 
\setlength\tabcolsep{10pt}
\resizebox{0.84\linewidth}{!} {
\begin{tabular}{l|ccc|ccc}
\toprule
\multirow{2}{*}{Method} & \multicolumn{3}{c|}{\tabincell{c}{Error (lower is better)}} & \multicolumn{3}{c}{\tabincell{c}{Accuracy (higher is better)}} \\\cline{2-7}
                                      & rel & log10 & rms & $\delta < 1.25$ & $\delta < 1.25^2$ & $\delta < 1.25^3$ \\\midrule
Top down structure       & 0.175  &  0.072 & 0.688 & 0.689 & 0.919 & 0.979 \\
Bottom up structure      & 0.169  & {0.071}& {0.673} & {0.698} & {0.923} & {0.981} \\
Skip connection structure      & 0.161  & 0.070& 0.664 & {0.709} & 0.923 & {0.981} \\
All to one structure     & \textbf{0.154}  & \textbf{0.068}& \textbf{0.648} & \textbf{0.725} & \textbf{0.927} & \textbf{0.981} \\
\bottomrule                            
\end{tabular}
}
\label{cnn_order}
\vspace{-8pt}
\end{table*}

\begin{figure*}[t]
\centering
\includegraphics[width=1\linewidth]{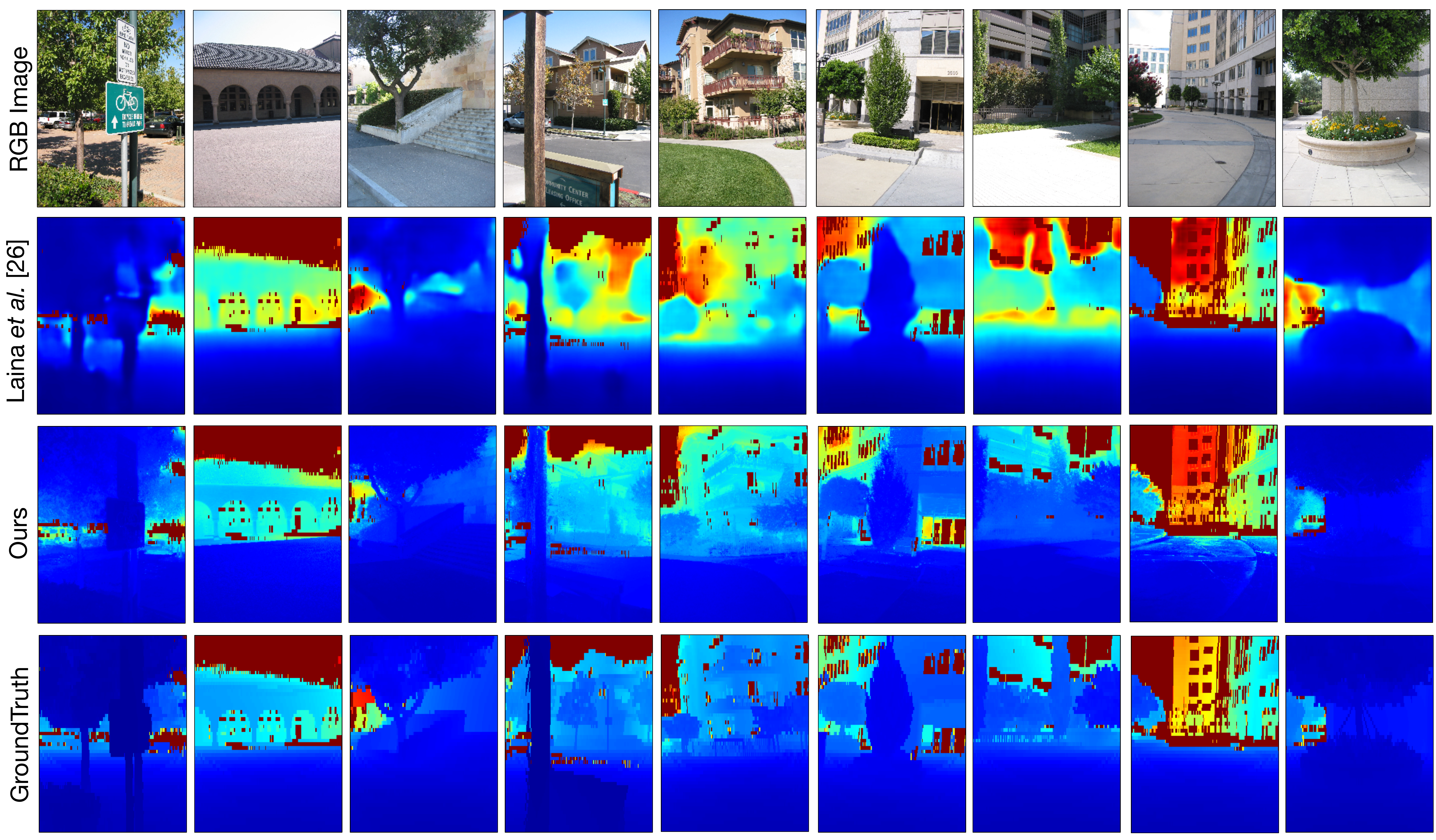} 
\vspace{-13pt}
\caption{Examples of depth prediction results on the Make3D dataset. The four rows from up to bottom are the input test RGB images, the results produced from Laina~\etal~\cite{laina2016deeper}, the results of our ResNet50-MSCRF model and the groundtruth depth maps, respectively.}
\label{make3d2}
\vspace{-10pt}
\end{figure*}


\subsubsection{Generation of multi-scale CNN side-outputs} 
Our approach can be applied with any multi-scale front-end CNN models including those with skip-connections. We here briefly describe the scheme we adopt to build CNN side outputs from the front-end CNN for the multi-scale fusion with CRFs. In~\cite{xie2015holistically} a convolutional layer is first used to generate a score map from the feature map and then a deconvolutional (\textit{deconv}) layer is adopted as a bilateral upsampling operator to enlarge the score map such as to obtain the same size of the input image. However, we noticed that by adopting the approach in~\cite{xie2015holistically} the generated side outputs associated to the feature maps with smaller size are very coarse, causing a lot scene details missing. To address this problem, after the convolutional layer, we stack several \textit{deconv} layers, each of them enlarging the output map by two times. A Rectified Linear Unit (ReLU) is applied after each \textit{deconv} layer. 
After the last deconv layer we use a crop layer to cut the extra margin and obtain a side output with the same resolution of the ground-truth image. We employ this scheme to obtain side outputs for AlexNet, VGG-16 and ResNet-50, while for VGG-CD and VGG-ED, we use the same setting as in~\cite{xie2015holistically}, as their decoder or deconvolutional part is able to obtain more fine-grained side outputs.  Table~\ref{tab:side-outputs} shows detailed network parameters used to obtain the side output from the last convolutional block of ResNet-50 (\ie from the layer \textit{res5c}).  

\subsubsection{Parameters settings}
As described in Section~\ref{sec:optimization}, training consists of a pretraining and a fine tuning phase. In the first phase, we train the front-end CNN with parameters initialized with the corresponding ImageNet pretrained models. For AlexNet, VGG-16, VGG-ED and VGG-CD, the batch size is set to 12 and for ResNet-50 to 8. 
The learning rate is initialized at $10^{-11}$ and decreases by 10 times around every 50 epochs. 80 epochs are performed for pretraining in total. 
The momentum and the weight decay are set to 0.9 and 0.0005, respectively. 
{When the pretraining is finished, we connect all the side outputs of the front-end CNN to our CRFs-based multi-scale deep models 
for end-to-end training of the whole network.} In this phase, the batch size is reduced to 6 and a fixed 
learning rate of $10^{-12}$ is used. The same parameters of the pre-training phase are 
used for momentum and weight decay. The bandwidth weights for the Gaussian kernels are obtained 
through cross validation. {The number of mean-field iterations is set to 5 for efficient training for both 
the cascade CRFs and multi-scale CRFs. We do not observe significant improvement using more than 5 iterations.} 
Training the whole network takes around $\sim 25$ hours on the Make3D dataset, $\sim 28$ hours on the KITTI dataset and $\sim 31$ hours on the NYU v2 dataset.

\begin{table*}
\centering
\caption{Overall performance comparison with state of the art methods on the NYU Depth V2 dataset. Our approach achieves the best on most of the metrics, while the runners-up Eigen and Fergus~\cite{eigen2015predicting} and  Laina \etal~\cite{laina2016deeper} employ more training data than ours. ResNet-50-unified means using ResNet-50 front-end network with the proposed multi-scale unified CRF model.}
\vspace{-3pt}
\setlength\tabcolsep{8pt}
\resizebox{0.92\linewidth}{!} {
\begin{tabular}{l|cccc|ccc}
\toprule
\multirow{2}{*}{Method} & \multicolumn{4}{c|}{\tabincell{c}{Error \\(lower is better)}} & \multicolumn{3}{c}{\tabincell{c}{Accuracy \\(higher is better)}} \\\cline{2-8}
                                      & rel & log10 & rms & rms (sc-inv.) & $\delta < 1.25$ & $\delta < 1.25^2$ & $\delta < 1.25^3$ \\\midrule
Karsch \etal~\cite{saxena2009make3d}                    & 0.349 &  -     & 1.214 & 0.325  &0.447 & 0.745 & 0.897 \\
Ladicky \etal~\cite{karsch2014depth}               &0.35    &0.131&1.20   & - &    -     &     -    &    -      \\
Liu \etal~\cite{liu2014discrete} & 0.335 & 0.127 & 1.06  & -  &    -     &     -    &     -     \\
Ladicky \etal~\cite{ladicky2014pulling}               &   -       &    -    & -   & -  & 0.542& 0.829&  0.941 \\
Zhuo \etal~\cite{zhuo2015indoor}                       & 0.305 &0.122& 1.04 & -   & 0.525& 0.838& 0.962 \\
Liu \etal~\cite{liu2015deep}                                 & 0.230 &0.095& 0.824 & -  & 0.614& 0.883&  0.975 \\
Wang \etal~\cite{wang2015towards}                   & 0.220 & 0.094&0.745 & - & 0.605 & 0.890 & 0.970 \\
Eigen \etal~\cite{eigen2014depth}                      & 0.215  &  -      & 0.907& 0.219  & 0.611 &  0.887  &  0.971\\
Roi and Todorovic~\cite{roymonocular}                                   & 0.187  &  0.078& 0.744 & - & - &  -  &  -\\
Eigen and Fergus~\cite{eigen2015predicting}               & 0.158  & -       & 0.641 & 0.171 & 0.769 & 0.950 & 0.988 \\
Laina \etal~\cite{laina2016deeper}                     & 0.129  &0.056& {0.583}& - &0.801 & 0.950 & 0.986\\
\midrule 
Ours (ResNet-50-unified-4.7K-bottom up)                                & 0.139  & 0.063 & 0.609 & 0.163 &  0.793 & 0.948 & 0.984 \\
Ours (ResNet-50-unified-95K-bottom up)                                 & {0.121}  & {0.052}& 0.586 & 0.149 & {0.811}  & {0.954} & {0.987} \\
Ours (ResNet-50-unified-95K-all to one)                                 & \textbf{0.108}  & \textbf{0.045}& \textbf{0.579} & \textbf{0.142} &\textbf{0.823}  & \textbf{0.957} &\textbf{0.987} \\
\bottomrule                           
\end{tabular}
}
\label{overall_nyu}
\vspace{-5pt}
\end{table*}


\subsection{Experimental Results}
To present the experimental results, we start from an ablation study for investigating the performance impact of different front-end network architectures, the effectiveness of the proposed CRF-based multi-scale fusion models and the influence of the stacking orders for making the sequential neural network. Then we compare the overall performance with the state of the art methods, and finally the qualitative results and running time are analyzed. 
\begin{table*}[!t]
\centering
\caption{Overall performance comparison with state of the art methods on the Make3D dataset. Our approach outperforms all the competitors w.r.t. the C2 Error, and performs only slightly worse on the \textit{rel} metric of the C1 Error than Laina \etal~\cite{laina2016deeper} using Huber loss and significantly larger training data.}
\vspace{-3pt}
\setlength\tabcolsep{8pt}
\resizebox{0.80\linewidth}{!} {
\begin{tabular}{l|cccc|ccc}
\toprule
\multirow{2}{*}{Method} & \multicolumn{4}{c|}{\tabincell{c}{C1 Error}} & \multicolumn{3}{c}{\tabincell{c}{C2 Error}} \\\cline{2-8}
                                      & rel & log10 & rms & rms (sc-inv.) & rel & log10 & rms \\
\midrule
Karsch \etal~\cite{karsch2014depth}           &0.355  & 0.127 & 9.20 & - & 0.361   & 0.148 & 15.10 \\
Liu et al.~\cite{liu2014discrete} & 0.335 & 0.137 & 9.49 & - & 0.338 & 0.134 & 12.60 \\
Liu et al.~\cite{liu2015deep}      & 0.314    &0.119 & 8.60 & -   & 0.307   &    0.125     &     12.89      \\
Li et al.~\cite{li2015depth}   &   0.278       &    0.092    &    7.19 & -    & 0.279      & 0.102& 10.27 \\
Laina \etal~\cite{laina2016deeper} ($\ell_2$ loss)    &   0.223 & 0.089 &  4.89 & -  & -       &    -     & - \\
Laina \etal~\cite{laina2016deeper} (Huber loss)    &   0.176 & 0.072 &  4.46 & -  & -       &    -     & -  \\
\midrule
Ours (ResNet-50-cascade-bottom up)          &   0.213  & 0.082 &   4.67 & 0.245  & 0.221  & 4.79 & 8.81 \\
Ours (ResNet-50-unified-bottom up) &   0.206 & 0.076 &   4.51  & 0.237 & 0.212 & 4.71 & 8.73 \\
Ours (ResNet-50-unified-10K-bottom up) &   0.184  &  {0.065}    &  {4.38}   & 0.219  & {0.198}  &   {4.53}  & {8.56} \\
Ours (ResNet-50-unified-10K-all to one) &   \textbf{0.174}  &   \textbf{0.059}    &   \textbf{4.27}   & \textbf{0.211}  & \textbf{0.185}  &   \textbf{4.41}  & \textbf{8.43} \\
\bottomrule                         
\end{tabular}
}
\label{overall_make3d}
\vspace{-12pt}
\end{table*}

\subsubsection{Evaluation of different front-end CNN architectures}
As discussed above, the proposed multi-scale CRF-based fusion models are 
general and different deep architectures can be used for the front-end network. In this section we evaluate the impact of this choice on the depth estimation performance. We consider both the case of the pretrained front-end models (\ie only side losses are employed but the multi-scale CRF models are not plugged), indicated with `pretrain', and the case of the fine-tuned models, including the front-end network with the multi-scale cascade CRFs (cascade-CRFs). The results of the experiments are shown in Table~\ref{cnn_arch}. As expected, in both cases deeper CNN architectures produced more accurate predictions, and ResNet-50 achieves the best performance among all the front-end networks. Moreover, VGG-CD is slightly better than VGG-ED, and both these models outperforms VGG-16, showing that the symmetric network structure is beneficial for the dense pixel-level prediction problems. Importantly, for all considered front-end networks there is a significant increase in performance when applying the proposed CRF-based models.

\par Figure~\ref{nyu2examples} depicts some examples of predicted depth maps using different front-end networks on the NYU Depth V2 test dataset. As we can see from the figure, the qualitative results confirm that the deeper architecture leads to better depth recovery. By comparing the reconstructed depth maps obtained with pretrained models (\eg using only the front-end networks VGG-CD and ResNet-50) with those generated with our multi-scale models, it is clear that our approach remarkably improves prediction accuracy and visual quality.

\subsubsection{Evaluation of different multi-scale CRF fusion models}
To evaluate the effectiveness of the proposed CRF-based multi-scale fusion models, we conduct experiments on the NYU Depth V2 dataset and consider the following baselines: 
\par(i) the `HED' method in~\cite{xie2015holistically}, where multiple side outputs are fused with a weighted averaging scheme and the sum of multiple side output losses is jointly minimized as deep supervision with a cross-entropy loss, while we use the square loss as our problem involves
continuous variables; 
\par(ii) the `Hypercolumn' method~\cite{hariharan2015hypercolumns}, where 
multi-scale feature maps generated from different semantic network layers are concatenated and fused;
\par (iii) a continuous CRF (`C-CRF') applied on the prediction of the front-end network, \ie plugging after the last output layer as a post-processing module without end-to-end training.
\par For the first two baselines, we want to compare our models with other popular methods for fusing multi-scale CNN information, while the third one aims at demonstrating the effectiveness of the continous CRF itself. In these experiments we consider VGG-CD as the front-end CNN architecture. The results of the comparison are shown in Table~\ref{comparison}. It is evident that with our CRF-based fusion models (both the cascade CRFs and the unified CRFs) more accurate depth maps can be obtained, demonstrating that our idea of integrating complementary information derived from CNN side output maps within a graphical model framework is more effective than traditional fusion schemes. 
Table~\ref{comparison} also compares the proposed cascade and unified models. As expected, the unified model produces more accurate depth maps, 
at the price of an increased computational cost. This can also be observed from Table~\ref{cnn_arch}. The C-CRF (in Table~\ref{comparison}) improves the depth estimation at all metrics over the VGG-CD (pretrain) (in Table~\ref{cnn_arch}) with a clear gap, showing the CRF model is very useful for refining the deeply predicted map. By jointly learning with the front-end (\ie end-to-end training), ours (single-scale) further boosts the performance. Finally, we analyze the impact of adopting multiple scales and compare our complete models (5 scales)
with their version when only a single and three side output layers are used. It is evident that the performance can be improved by increasing the
number of scales. 

\begin{figure*}[t]
\centering
\includegraphics[width=1\linewidth]{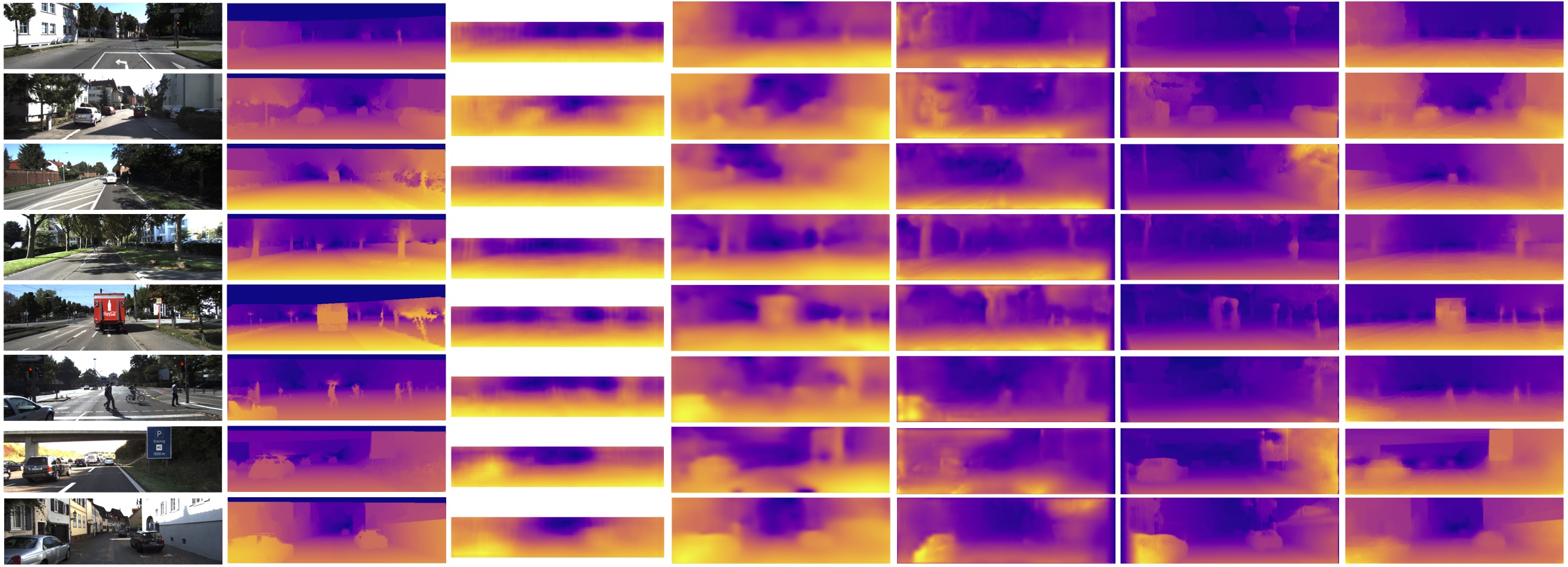} 
\put(-499,191){\footnotesize RGB Image}
\put(-434,191){\footnotesize GT Depth Map}
\put(-364,191){\footnotesize Eigen~\etal~\cite{eigen2014depth}}
\put(-285,191){\footnotesize Zhou~\etal~\cite{zhou2017unsupervised}}
\put(-211,191){\footnotesize Garg~\etal~\cite{garg2016unsupervised}}
\put(-143,191){\footnotesize Godard~\etal~\cite{godard2016unsupervised}}
\put(-51,191){\footnotesize Ours}
\caption{Examples of depth prediction results on the KITTI raw dataset. Qualitative comparison with other depth estimation methods on this dataset is presented. The sparse ground-truth depth maps are interpolated for better visualization.}
\label{KITTI_comp}
\vspace{-10pt}
\end{figure*}
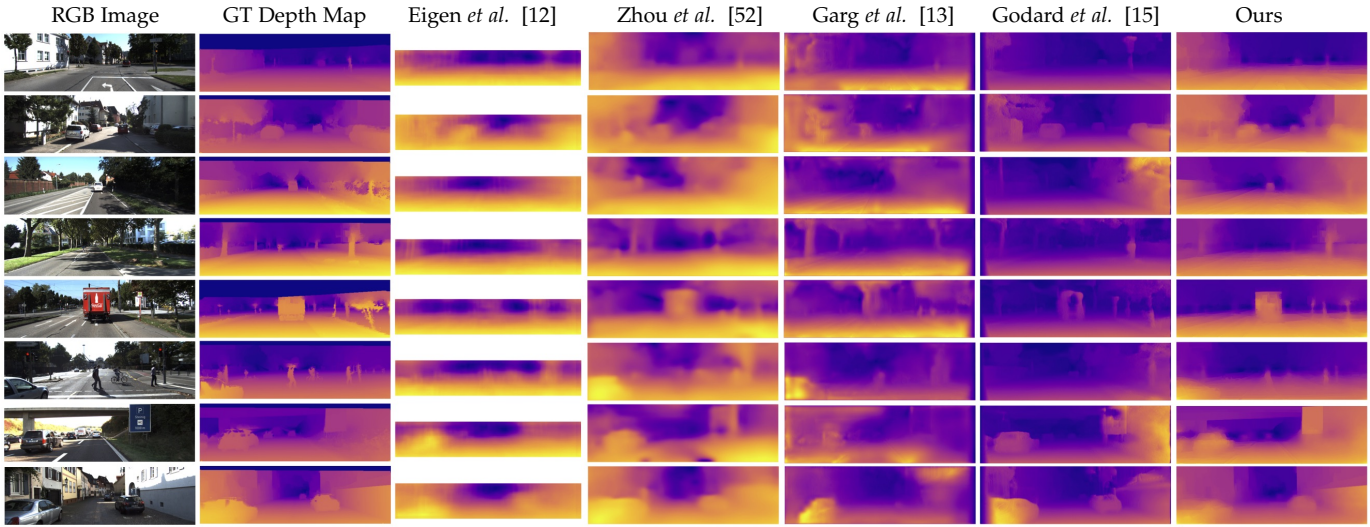

\begin{table*}
\centering
\caption{Overall performance comparison with state of the art methods on the KITTI raw dataset. Our approach obtains very competitive performance over all the competitors w.r.t. all the evaluation metrics on the testing set given by Eigen~\etal~\cite{eigen2014depth}. For the setting, caps means different gt/predicted depth range and stereo means using left and right images captured from two monocular cameras in the training phase. Ours uses a unified model considering both the bottom up and the all to one network structure.}
\vspace{-3pt}
\setlength\tabcolsep{8pt}
\resizebox{0.97\linewidth}{!} {
\begin{tabular}{l|cc|cccc|ccc}
\toprule
\multirow{2}{*}{Method} & \multicolumn{2}{c|}{Setting} & \multicolumn{4}{c|}{\tabincell{c}{Error (lower is better)}} & \multicolumn{3}{c}{\tabincell{c}{Accuracy (higher is better)}} \\\cline{2-10}
                                      & range & stereo & rel & sq rel & rms & rms (sc-inv.) & $\delta < 1.25$ & $\delta < 1.25^2$ & $\delta < 1.25^3$ \\\midrule
Saxena \etal~\cite{saxena2009make3d}   & 0-80m & No & 0.280  &  - & 8.734 & 0.327 & 0.601  &  0.820  &  0.926\\
Eigen \etal~\cite{eigen2014depth}    & 0-80m & No & 0.190  &  -      &  7.156 & 0.246  &  0.692 &  0.899  &  0.967 \\
Liu \etal~\cite{liu2015deep}    & 0-80m & No & 0.217  &  0.092   &  7.046 & - &  0.656 &  0.881  &  0.958 \\
Zhou~\etal~\cite{zhou2017unsupervised} & 0-80m & No  & 0.208   &  1.768  & 6.858  & - & 0.678   & 0.885   & 0.957  \\ 
Kuznietsov~\etal \cite{kuznietsov2017semi} (only supervised) & 0-80m & No & - & - & 4.815 & - & {0.845} & {0.957} & {0.987} \\ \midrule
Garg~\etal~\cite{garg2016unsupervised}  & 0-80m & Yes & 0.177 & 1.169  & 5.285   & -  &  0.727 & 0.896  &  0.962   \\
Garg~\etal~\cite{garg2016unsupervised} L12 + Aug 8x  & 1-50m & Yes & 0.169  & 1.080   & 5.104  & -  &  0.740 & 0.904  &  0.958   \\
Godard \etal~\cite{godard2016unsupervised}  & 0-80m & Yes & 0.148 & 1.344  & 5.927 & -   &  0.803 & 0.922  &  0.964   \\
Kuznietsov \etal \cite{kuznietsov2017semi}  & 0-80m & Yes & - & - & \textbf{4.621} & -  & \textbf{0.852} & \textbf{0.960} & \textbf{0.986}  \\
\midrule
Ours (ResNet-50 Pretrain)  &0-80m & No & 0.152  & 0.973  & 4.902  & 0.176 & 0.782 & 0.931 & 0.975 \\
Ours (ResNet-50 Fine-tune-bottom up) & 0-80m & No & 0.132   & 0.911 & 4.791 & 0.162  & 0.804  & 0.945 &  0.981\\
Ours (ResNet-50 Fine-tune-all to one) & 0-80m & No & \textbf{0.125}  & \textbf{0.899} & {4.685} & \textbf{0.154} & 0.816  & {0.951} & {0.983} \\
\bottomrule                           
\end{tabular}
}
\label{overall_KITTI}
\vspace{-10pt}
\end{table*}

\subsubsection{Evaluation of multi-scale message passing structures}
We evaluate the influence of different multi-scale message passing structures using the cascade CRF model. Four connection structures as depicted in Fig.~\ref{fig:messagepassingstructure} are compared. Table~\ref{cnn_order} shows the monocular depth estimation results on NYUD-v2 dataset. The comparison results confirm that the message passing structure indeed has an impact on the final performance. The bottom up and top down structures have similar performance, while the skip-connection structure slightly outperform these two. The all to one structure performs the best, producing around 2.0\% gain in terms of the \textit{rel} metric than the top down structure, which means that directly passing message to the finest prediction scale from the rest scales can absorb more complementary information than the gradual passing fashions used in the first three structures. 

\vspace{-5pt}
\subsubsection{Comparison with state of the art}
We also compare our approach with state of the art methods on all the datasets.
For previous works we directly report results taken from the original papers. Table~\ref{overall_nyu} shows 
the results of the comparison on the NYU Depth V2 dataset. 
For our approach we consider the cascade model and use two different training sets for pretraining: the small set of
4.7K pairs employed in all our experiments and a larger set of 95K images as in \cite{laina2016deeper}. Note that for fine tuning
we only use the small set. As shown in the table, our approach outperforms all competing methods 
and it is the second best model when we use only 4.7K images. This is remarkable considering that,
for instance, in~\cite{eigen2015predicting} 120K image pairs are used for training. Our model achieves the best results on all the metrics via using 95K pretraining samples and using the proposed all to one message passing structure.

We also perform a comparison with several state of the art methods on the Make3D dataset (Table~\ref{overall_make3d}).
Following~\cite{liu2014discrete}, the error metrics are computed in two different settings, \ie considering (C1) only 
the regions with ground-truth depth less than 70 and (C2) the entire image. 
It is clear that the proposed approach is significantly better than previous methods. 
In particular, comparing with Laina~\etal~\cite{laina2016deeper}, the best performing method in the literature, 
it is evident that our approach, both in case of the cascade and the multi-scale models, outperforms \cite{laina2016deeper} by a 
significant margin when Laina \etal also adopt a square loss. It is worth noting that in \cite{laina2016deeper} a 
training set of 15K image pairs is considered, while we employ much less training samples. 
By increasing our training data (\ie $\sim 10$K in the pretraining phase), our multi-scale CRF model also outperforms \cite{laina2016deeper}
with Huber loss (log10 and rms metrics). The final performance is further boosted by considering the all to one structure similar to NYUD v2 dataset. Finally, it is very interesting to compare the proposed method with the 
approach in Liu \etal\cite{liu2015deep}, since in \cite{liu2015deep} a CRF model is also employed within a deep network trained end-to-end. 
Our method significantly outperforms \cite{liu2015deep} in terms of accuracy. Moreover, in \cite{liu2015deep} a time of 1.1sec is reported 
for performing inference on a test image but the time required by superpixels calculations is not taken into account. Oppositely, with our method computing the depth map for a single image takes about 1 sec in total.
\par The state of the art comparison on KITTI dataset is shown in Table~\ref{overall_KITTI}. The competitors include Saxena \etal~\cite{saxena2005learning}, Eigen \etal~\cite{eigen2014depth}, Liu \etal~\cite{liu2016learning}, Zhou~\etal~\cite{zhou2017unsupervised}, Garg~\etal~\cite{garg2016unsupervised}, Godard~\etal~\cite{godard2016unsupervised} and Kuznietsov \etal \cite{kuznietsov2017semi}. As the same setting of ours, the first four methods use single monocular images in the training phase, while the last two considered two monocular images with a stereo setting for training. Among the first four competitors, Eigen \etal~\cite{eigen2014depth} significantly outperforms the others in terms of the metric of the mean relative error (\textit{rel}), due to the usage of large-scale training data (more than 1 million samples). While our model achieves much better performance than Eigen~\etal~\cite{eigen2014depth} in all metrics with much less data (22.6K samples). Although the training of the last two methods (requiring two monocular images) is not equal to our setting, the proposed approach with both the bottom-up and the all to one structures still produces better results than them with clear performance gap in all metrics. Kuznietsov \etal \cite{kuznietsov2017semi} reports results for both the stereo training and the monocular supervised training. It is not directly comparable with the stereo training setting, which is significantly different as it requires both left and right images from a binocular camera. Ours focuses on monocular depth estimation and achieves lower error performance comparing with theirs using the same monocular setting. Fig.~\ref{KITTI_comp} also shows some qualitative comparison results with these methods, further demonstrating the advantageous performance of our approach.

\begin{figure}[t]
\centering
\includegraphics[width=.98\linewidth]{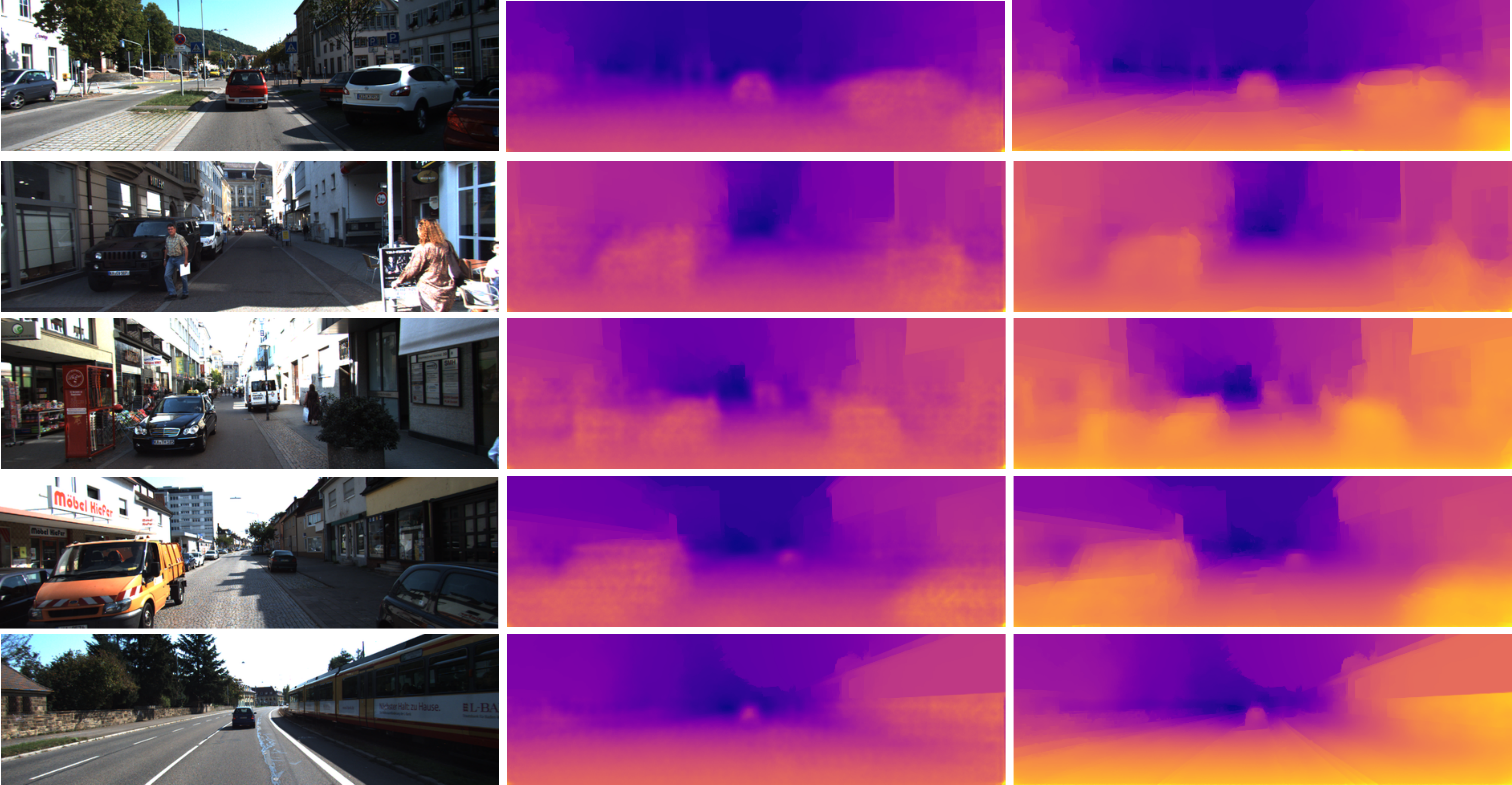} 
\caption{Examples of depth prediction results on the KITTI raw dataset. The middle column and the right column show the pretrained and the fine-tuned estimation results respectively.}
\label{KITTI_results}
\vspace{-10pt}
\end{figure}

\subsubsection{Qualitative depth estimation results}
Fig.~\ref{nyu2examples},~\ref{make3d2} and~\ref{KITTI_results} show some examples of the qualitative depth estimation results and the comparison with the competing methods on the NYUD-V2, Make3D and KITTI dataset respectively. It is clear that the proposed approach is able to produce sharper depth estimation with better visual quality compared with the classic CNN structures, which demonstrates the importance of the prediction aided by the CRFs with appearance and smoothness constraints. Fig.~\ref{KITTI_results} also shows a qualitative comparison between the pretrained front-end CNN and the fine-tuned whole model. It can be observed that our approach can recover more scene structures and details. We believe that this is probably because the effective structured fusion of the coarse-to-fine multi-scale predictions of the deep network with the proposed CRF models. For the influence of the variance in the CRF model on the prediction errors, as the variance term is actually acted as a normalization factor after the message passing. It may have influence but the main influence is dominated by the predictions of deep front-end CNN based on our observation from the experimental results.
\subsubsection{Empirical run-time analysis}
Computational run-time complexity is an important aspect for deep structured prediction models. In this paragraph we provide a short discussion about the computational cost of the proposed CRFs-based models. As shown in the paper, the multi-scale CRF model achieves better accuracy and lower error than the cascade model for both the NYU Depth V2 and the Make3D experiments. However, as expected, the cascade model is more advantageous in terms of the running time. For instance, considering ResNet-50 as the front-end CNN, the time required at test phase for one image is $1.02$ seconds w.r.t. the cascade model and $1.45$ seconds w.r.t. the multi-scale model, and the image resolution is $320 \times 240$ pixels. Higher resolution of the network input usually brings more computational overhead. We also test the running time given the input resolution of $640\times 480$ and it costs around $2.25$ seconds for processing one image. We believe that if we reduce the receptive field of the CRF model from fully connected to partially connected, the computing time could be significantly reduced. 

\section{Conclusion}\label{sec:conclusion}
In this paper, we introduced a novel approach for predicting depth maps from a single RGB image. The core of the method is a novel framework based on continuous CRFs for fusing multi-scale score-level side-outputs derived from different semantic CNN layers. We demonstrated that this framework can be used in combination with several common CNN architectures and can be implemented for end-to-end training. The extensive experiments confirmed the validity of the proposed multi-scale fusion approach. While this paper specifically addresses the problem of depth prediction, we believe that other tasks in computer vision involving pixel-level predictions of continuous variables, can also benefit from our implementation of the mean-field updating within the CNN framework.

Currently, the multi-scale fusion is performed on the score level. Further research direction will investigate the integration of both the feature- and the score-level multi-scale information within a unified graphical model. Moreover, the study of strategies for further improving the training and testing efficiency of the CNN-CRF models will also be an interesting aspect in the future work. The monocular depth estimation is particularly useful for various cross-modal recognition
and detection
tasks. A straightforward follow-up of this work would be designing a joint multi-task deep model to transfer the learned depth model for aiding other similar dense prediction problems~\cite{xu2018PAD-Net} such as contour detection and semantic segmentation.

\vspace{-5pt}
{\small
\bibliographystyle{ieee}
\bibliography{egbib}
}
\vspace{-15pt}
%

\begin{IEEEbiography}
{Dan Xu}
 is a Ph.D. candidate in the Department of Information Engineering and Computer Science, and a member of Multimedia and Human Understanding Group (MHUG) led by Prof. Nicu Sebe at the University of Trento. He was a research assistant in the Multimedia Laboratory in the Department of Electronic Engineering at the Chinese University of Hong Kong. 
His research focuses on computer vision, multimedia and machine learning. Specifically, he is interested in deep learning, structured prediction and cross-modal representation learning and the applications to scene understanding tasks. He received the Intel best scientific paper award at ICPR 2016.
\end{IEEEbiography}

\begin{IEEEbiography}
{Elisa Ricci}
received the PhD degree from the
University of Perugia in 2008. She is an assistant
professor at the University of Perugia and a
researcher at Fondazione Bruno Kessler. She
has since been a post-doctoral researcher at
Idiap, Martigny, and Fondazione Bruno Kessler,
Trento. She was also a visiting researcher at the
University of Bristol. Her research interests are
mainly in the areas of computer vision and
machine learning. She is a member of the IEEE.
\end{IEEEbiography}

\begin{IEEEbiography}
{Wanli Ouyang}
received the PhD degree in the
Department of Electronic Engineering, The Chinese
University of Hong Kong. He is now
a senior lecturer in the School of Electrical and Information Engineering at the University of Sydney, Australia. His research interests include image processing, computer vision and pattern recognition. He is a senior member of IEEE.
\end{IEEEbiography}

\begin{IEEEbiography}
{Xiaogang Wang}
received the PhD degree in Computer Science from Massachusetts Institute of Technology. He is an associate professor in the Department of Electronic Engineering at the Chinese University of Hong Kong since August 2009. He was the Area Chairs of ICCV 2011 and 2015, ECCV 2014 and 2016, ACCV 2014 and 2016. He received the Outstanding Young Researcher in Automatic Human Behaviour Analysis Award in 2011, Hong Kong RGC Early Career Award in 2012, and CUHK Young Researcher Award 2012.
\end{IEEEbiography}


\begin{IEEEbiography}
{Nicu Sebe} is Professor with the University of Trento,
Italy, leading the research in the areas of multimedia
information retrieval and human behavior
understanding. He was the General Co- Chair of the
IEEE FG Conference 2008 and ACM Multimedia
2013, and the Program Chair of the International
Conference on Image and Video Retrieval in 2007
and 2010, ACM Multimedia 2007 and 2011. He was
the Program Chair of ICCV 2017 and ECCV 2016,
and a General Chair of ACM ICMR 2017. He is a fellow of the International Association
for Pattern Recognition.
\end{IEEEbiography}




\end{document}